%% file: neurips_2025.tex
\documentclass{article}

\usepackage[preprint]{neurips_2025}

\usepackage{colortbl}
\usepackage[most]{tcolorbox}
\usepackage{CJKutf8}
\newcommand{\cc}[1]{\begin{CJK*}{UTF8}{gbsn}#1\end{CJK*}}
\usepackage[utf8]{inputenc} 
\usepackage[T1]{fontenc}    
\usepackage{hyperref}       
\usepackage{url}            
\usepackage{booktabs}       
\usepackage{amsfonts}       
\usepackage{nicefrac}       
\usepackage{microtype}      
\usepackage{xcolor}         
\usepackage{wrapfig}
\usepackage{graphicx}

\usepackage{amsmath,amssymb,amsfonts}
\usepackage{graphicx}
\usepackage{textcomp}

\usepackage{longtable}
\usepackage{booktabs}

\usepackage{algorithm}
\usepackage{algorithmicx}
\usepackage{algpseudocode}
\usepackage{tikz}
\usepackage{pgf-pie}   
\usepackage{caption}   
\usepackage{amsmath}
\usepackage{amsfonts,amssymb}

\usepackage{mathrsfs}
\usepackage{multirow}
\usepackage{makecell}

\usepackage{pgfplots}
\usepackage{listings}
\usepackage{enumitem}
\usepackage{listings}
\usepackage{amsmath,amsfonts}
\usepackage{array}
\usepackage[caption=false,font=normalsize,labelfont=sf,textfont=sf]{subfig}
\usepackage{textcomp}
\usepackage{url}
\usepackage{verbatim}
\usepackage{graphicx}
\usepackage{pgf-pie}
\usepackage{amsmath}
\usepackage{booktabs}
\usepackage{makecell}
\usepackage{tikz}
\usepgfplotslibrary{groupplots}
\pgfplotsset{compat=1.17} 
\usepackage{etoolbox}

\usepackage{tabularx}
\usepackage{booktabs}

\title{OneEval: Benchmarking LLM Knowledge-intensive Reasoning over Diverse  Knowledge Bases}

\author{%
Yongrui Chen\textsuperscript{1,2}, 
Zhiqiang Liu\textsuperscript{3}, 
Jing Yu\textsuperscript{3}, 
Lin Ren\textsuperscript{1,2}, 
Nan Hu\textsuperscript{1,2}, 
Xinbang Dai\textsuperscript{1,2},\\ 
\textbf{Jiajun Liu\textsuperscript{1,2}, 
\textbf{Jiazhen Kang}\textsuperscript{1,2},
\textbf{Shenyu Zhang}\textsuperscript{1,2}, 
\textbf{Xinda Wang}\textsuperscript{3}, 
\textbf{Keyan Ding}\textsuperscript{3}, 
\textbf{Pengfei Shen}\textsuperscript{4},} \\
\textbf{Haolei Zhu\textsuperscript{4}, 
\textbf{Hongjie Deng}\textsuperscript{3}, 
\textbf{Yisong Wang}\textsuperscript{5}, 
\textbf{Tongtong Wu}\textsuperscript{6}, 
\textbf{Sheng Bi}\textsuperscript{1}, 
\textbf{Wen Zhang}\textsuperscript{3},}\\
\textbf{Tianxing Wu\textsuperscript{1,2}, 
\textbf{Qiu Ji}\textsuperscript{4}, 
\textbf{Haofen Wang}\textsuperscript{5}, 
\textbf{Wenliang Chen}\textsuperscript{7}, 
\textbf{Huajun Chen}\textsuperscript{3}, 
\textbf{Guilin Qi}\textsuperscript{1, 2}}\\
\textsuperscript{1}Southeast University, China \\
\textsuperscript{2}Key Laboratory of New Generation Artificial Intelligence Technology and its Interdisciplinary \\
Applications (Southeast University), Ministry of Education, China \\
\textsuperscript{3}Zhejiang University, China \\
\textsuperscript{4}Nanjing University of Posts and Telecommunications, China \\
\textsuperscript{5}Tongji University, China \\
\textsuperscript{6}Monash University, Australia \\
\textsuperscript{7}Soochow University, China \\
\texttt{\{yongruichen, gqi\}@seu.edu.cn}
}

\begin{document}

\maketitle

\begin{abstract}
Large Language Models (LLMs) have demonstrated substantial progress on reasoning tasks involving unstructured text, yet their capabilities significantly deteriorate when reasoning requires integrating structured external knowledge such as knowledge graphs, code snippets, or formal logic. This limitation is partly due to the absence of benchmarks capable of systematically evaluating LLM performance across diverse structured knowledge modalities. To address this gap, we introduce \textbf{\textsc{OneEval}}, a comprehensive benchmark explicitly designed to assess the knowledge-intensive reasoning capabilities of LLMs across four structured knowledge modalities—unstructured text, knowledge graphs, code, and formal logic—and five critical domains (general knowledge, government, science, law, and programming). \textsc{OneEval} comprises 4,019 carefully curated instances and includes a challenging subset, \textsc{OneEval}\textsubscript{Hard}, consisting of 1,285 particularly difficult cases. Through extensive evaluation of 18 state-of-the-art open-source and proprietary LLMs, we establish three core findings: a) \emph{persistent limitations in structured reasoning}, with even the strongest model achieving only 32.2\% accuracy on \textsc{OneEval}\textsubscript{Hard}; b) \emph{performance consistently declines as the structural complexity of the knowledge base increases}, with accuracy dropping sharply from 53\% (textual reasoning) to 25\% (formal logic); and c) \emph{diminishing returns from extended reasoning chains}, highlighting the critical need for models to adapt reasoning depth appropriately to task complexity. We release the \textsc{OneEval} datasets, evaluation scripts, and baseline results publicly, accompanied by a leaderboard to facilitate ongoing advancements in structured knowledge reasoning.

\end{abstract}

\section{Introduction}

Large Language Models (LLMs)~\cite{gpt4o,o1,deepseekv3} have recently demonstrated significant progress in complex reasoning tasks. Capabilities like synthesizing reasoning paths from unstructured text and leveraging implicit commonsense knowledge have advanced considerably~\cite{ott2023thoughtsource, wei2022chain}, as evidenced by performance on various benchmarks~\cite{mmlu,agieval,mtbench,openassistant}. However, even for models explicitly optimized for reasoning, LLMs continue to exhibit significant fragility and inaccuracy when required to integrate structured external knowledge bases such as code, knowledge graphs, or formal logic~\cite{bubeck2023sparks}.

A critical limitation underlying this challenge is the nature of existing reasoning benchmarks. They predominantly focus on unstructured textual reasoning \cite{hotpotqa, strategyqa}, which primarily involves synthesizing information from narrative text. In contrast, many practical reasoning tasks necessitate the processing and integration of structured knowledge formats such as codes, KGs, or formal logical statements. These formats differ fundamentally from the unstructured text that constitutes the majority of LLM training data. Consequently, the performance landscape of state-of-the-art LLMs—including powerful models like DeepSeek R1~\cite{deepseekr1}, Grok3~\cite{grok3}, and o3~\cite{o3o4}—across diverse knowledge modalities (e.g., formal logic, symbolic manipulation), and how performance trends change when transitioning from familiar textual distributions to these structured forms \cite{bubeck2023sparks}, remains inadequately characterized.

\begin{figure*}
\centering
	\includegraphics[width=\textwidth]{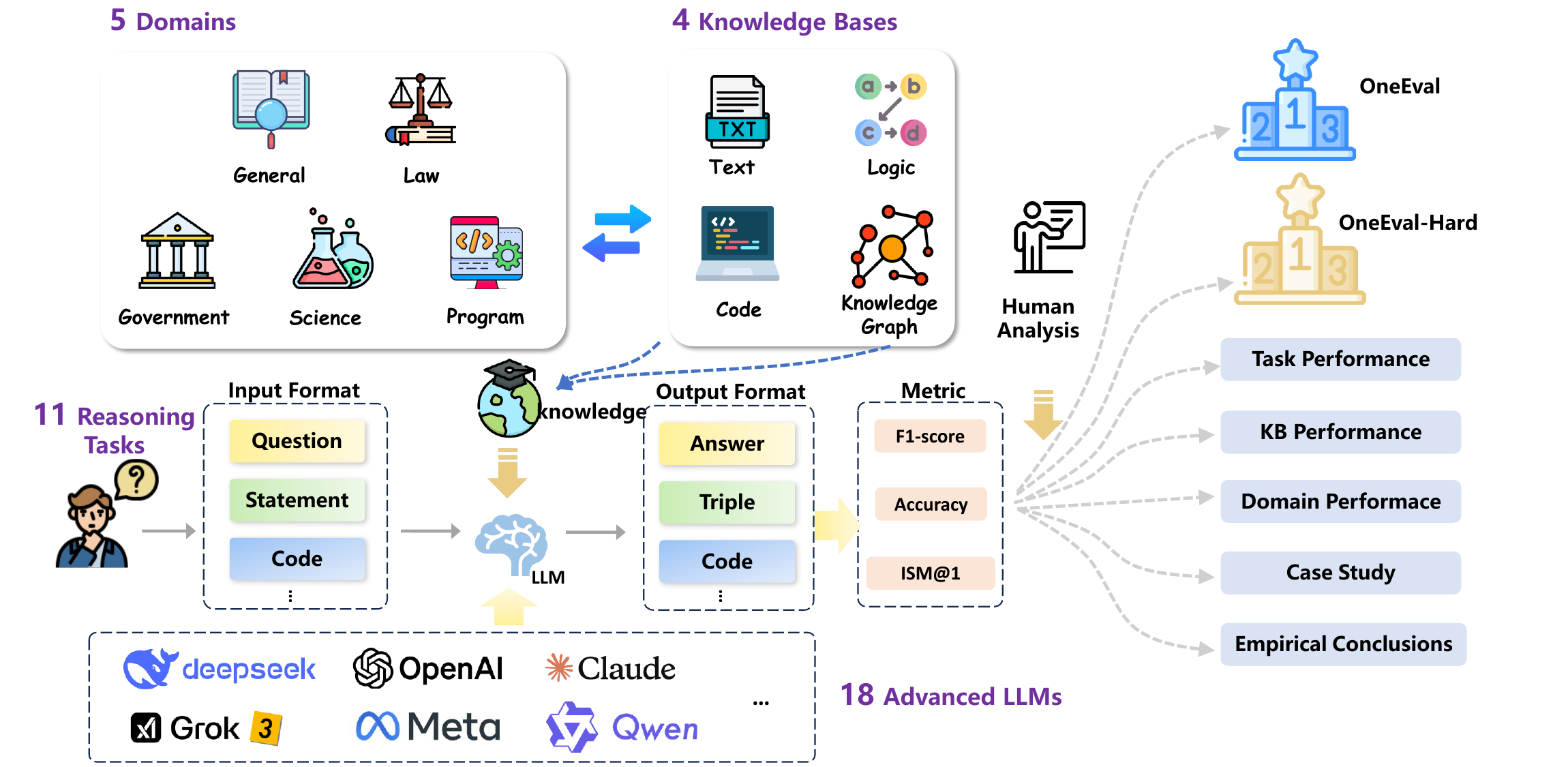}
	\caption{Overall framework of \textsc{OneEval}, covering 4 knowledge bases, 5 fields, and 11 tasks.} \label{fig:overview}
\end{figure*}

Motivated by this critical need for a comprehensive evaluation framework, we introduce \textbf{\textsc{OneEval}}, the first unified benchmark specifically designed to assess LLM reasoning capabilities across a spectrum of knowledge modalities. As illustrated in Figure~\ref{fig:overview}\textsc{OneEval} covers unstructured text, knowledge graphs, code, and formal logic, spanning five high-impact domains (general knowledge, government, science, law, and programming). Comprising 4,019 instances meticulously curated from 11 meticulously curated dataset, the benchmark also includes a \textsc{OneEval}\textsubscript{Hard} subset of 1,285 particularly challenging cases. Each instance provides a query paired with an external knowledge base presented in one of the specified modalities, alongside a gold answer.

A large-scale evaluation campaign involving 18 diverse open-source and proprietary models (Table~\ref{tab:overall_results}) yields three robust findings concerning the current state of LLM reasoning on structured knowledge:
a) \textit{Current LLMs remain significantly challenged by knowledge-intensive reasoning}: Even the most capable model achieves merely 32.2\% accuracy on the challenging \textsc{OneEval}\textsubscript{Hard} subset, indicating substantial room for improvement in reliability.
b) \textit{Reasoning performance degrades consistently as the knowledge modality becomes more structured}: Average accuracy declines sharply from 53\% on unstructured text to 25\% on formal logic, revealing a clear dependency on the input structure.
c) \textit{Longer reasoning chains offer diminishing, and eventually negative, returns}: Beyond a moderate length, the introduction of noise and potential errors outweighs the benefits of additional steps, highlighting the need for chain-length sensitivity in complex reasoning across modalities. Our key contributions include:

\begin{itemize}[leftmargin=1em]
    \item \textbf{Benchmark.} We release \textsc{OneEval}, the first comprehensive LLM knowledge-intensive reasoning benchmark spanning four types of knowledge bases, five domains, 4,019 instances, and a 1,285-sample hard subset, together with unified evaluation scripts.
    \item \textbf{Analysis.} We present a comprehensive and detailed study encompassing 18 advanced LLMs. Our work reveals crucial insights into the interplay between factors such as the degree of knowledge structuring, model size, and response length, and their impact on LLM reasoning performance.
    \item \textbf{Resources.} All datasets, prompts, and backbone LLM outputs are publicly available, and an online leaderboard encourages rapid community progress on knowledge-intensive reasoning.
\end{itemize}

\section{\textsc{OneEval}}

We introduce \textsc{OneEval}, a comprehensive benchmark designed to rigorously evaluate the knowledge reasoning capabilities of LLMs across a spectrum of knowledge base (KB) types and reasoning challenges. \textsc{OneEval} is constructed from 15 distinct public datasets, carefully selected to represent diverse facets of knowledge, including structured, unstructured, explicit, and implicit forms. To create a manageable yet representative evaluation set, for each constituent dataset $\mathcal{D}^i$ (where $i \in \{1, \dots, 15\}$), we randomly sample $\mathcal{N}^i$ instances. The final \textsc{OneEval} benchmark comprises the union of these sampled instances. Detailed statistics of the selected datasets, including their sources and sampling sizes $\mathcal{N}^i$, are provided in Table~\ref{tab:dataset}. Further specifics regarding the source datasets and their original task formats can be found in the Appendix.

\subsection{Task Definition}
Given a user query $\mathcal{Q}$ and an accessible knowledge base $\mathcal{S}$, the objective is to generate the desired answer $\mathcal{A}$ by leveraging information available within $\mathcal{S}$. Formally, the task for an LLM $f_\theta$ is to compute $\mathcal{A} = f_\theta(\mathcal{Q}, \mathcal{S})$. Here, the query $\mathcal{Q}$ can be presented in various formats, including natural language questions, statements, descriptions, or code snippets. The knowledge base $\mathcal{S}$ is drawn from a predefined set of distinct types, which we detail in the following subsection. The answer $\mathcal{A}$ should be a valid derivation or inference based on the provided $\mathcal{Q}$ and $\mathcal{S}$, and its format can vary, encompassing free-form text, structured outputs like triples, boolean values, or code snippets.

\subsection{Knowledge Base Types}

A critical aspect distinguishing \textsc{OneEval} is its focus on evaluating LLM performance conditioned on diverse types of external knowledge bases. This allows for a nuanced analysis of their capabilities across different data structures and reasoning paradigms. We categorize the knowledge bases in \textsc{OneEval} into five principal types, forming the set $\Omega_{\text{KB}} = \{\mathcal{D}, \mathcal{T}, \mathcal{K}, \mathcal{C}, \mathcal{L}\}$, where:

\begin{itemize}[leftmargin=1em]
    \item \textbf{Textual Base ($\mathcal{D}$):} A textual base $\mathcal{D} = \{d_1, d_2, \dots, d_n\}$ represents an unstructured collection of natural language documents or passages, where each $d_i$ is a unit of text.


    \item \textbf{Knowledge Graph ($\mathcal{K}$):} A knowledge graph (KG) $\mathcal{K}$ is a structured semantic network typically defined as a set of triples $\mathcal{K}=\{\left \langle s, p, o\right \rangle | s \in \mathcal{E}, p \in \mathcal{R}, o \in \mathcal{E} \cup \mathcal{L} \}$, where $\mathcal{E}$ is the set of entities, $\mathcal{R}$ is the set of relations, and $\mathcal{L}$ is the set of literal values. 
    
    \item \textbf{Code Bases ($\mathcal{C}$):} A code base $\mathcal{C} = \{c_1, c_2, \dots, c_m\}$ represents a collection of programmatic knowledge artifacts. Each $c_i$ is an element such as source code snippets, function definitions, API documentation entries, or specifications. 
    
    \item \textbf{Logic Bases ($\mathcal{L}$):} A logic base $\mathcal{L}$ provides a formal, explicit specification of a domain's conceptualization. It is typically defined by a tuple $\mathcal{L} = (\mathcal{C}_{ont}, \mathcal{P}_{ont}, \mathcal{A}_{ont})$, where $\mathcal{C}_{ont}$ is the set of concepts (classes), $\mathcal{P}_{ont}$ is the set of properties (relations between concepts), and $\mathcal{A}_{ont}$ is a set of axioms or rules defining constraints and logical relationships. 
\end{itemize}

\begin{table*} 
	\begin{center}
	{\caption{Statistics of each dataset in \textsc{OneEval}.}\label{tab:dataset}}
	\scalebox{0.85}{
		\begin{tabular}{clcccccc}
			\toprule
			\textbf{Knowledge Base} & \textbf{Task}  & \textbf{Input} & \textbf{Output} & \textbf{Language} & \textbf{\# of Samples} & \textbf{Metric} \\
            
			\cmidrule(lr){1-1} \cmidrule(lr){2-2} \cmidrule(lr){3-7} 
            \multirow{6}[1]{*}{\makecell[c]{Text}}
             & BioTextQA  & Question & Answer & English & 210 & F1 \\ 
             & MatTextQA  & Question & Answer & English & 210 & F1 \\ 
             & ChineseLawFact  & Description & Boolean & Chinese & 800 & Acc \\
             & AttributionNLI & Question & Answer & English & 210 & F1  \\
             & KCQAD & Question & Answer & English & 500 & F1  \\
             \cmidrule(lr){1-1} \cmidrule(lr){2-2} \cmidrule(lr){3-7} 
            \multirow{4}[1]{*}{\makecell[c]{Knowledge \\Graph}}
             & PharmKGQA  & Question & Answer & English & 210 & F1 \\ 
             & AffairQA  & Question & Answer & Chinese & 200 & Acc \\ 
             & PeopleRelQA  & Question & Answer & Chinese & 200 & F1 \\ 
             & ReportFixer  & Description & Triple & Chinese & 200 & F1 \\ 
             \cmidrule(lr){1-1} \cmidrule(lr){2-2} \cmidrule(lr){3-7} 
             \cmidrule(lr){1-1} \cmidrule(lr){2-2} \cmidrule(lr){3-7} 
            \multirow{1}[1]{*}{\makecell[c]{Code}}
            & VersiCode & Code & Code & English & 739 & ISM@1 \\ 
            \cmidrule(lr){1-1} \cmidrule(lr){2-2} \cmidrule(lr){3-7} 
            \multirow{1}[1]{*}{\makecell[c]{Logic}}
            & SymTex-ASC & Statement & Code & English & 540 & EM \\ 
            \bottomrule
		\end{tabular}
		}
	\end{center}
\end{table*}

\subsection{Domain}
\textsc{OneEval} spans five key knowledge domains, General, Government Affairs, Science, Law, and Programming, with a strong emphasis on the breadth and specialization of multi-source heterogeneous knowledge. It aims to systematically evaluate LLMs in terms of reasoning and application capabilities in complex, knowledge-driven tasks. The specific domain categories are illustrated in Appendix.

\subsection{\textsc{OneEval}-Hard}

To enable a more accurate and fine-grained evaluation of LLM performance, particularly in high-difficulty reasoning scenarios, we have manually curated a challenging subset of \textsc{OneEval}, denoted as \textsc{OneEval}-Hard. The construction of \textsc{OneEval}-Hard involved a multi-stage process combining empirical performance analysis with expert qualitative review.

For each test sample $(\mathcal{Q}, \mathcal{S}, \mathcal{A})$ from the full \textsc{OneEval}, we first empirically assessed its difficulty by evaluating a set of $K$ LLMs, $\mathcal{F} = \{f_\theta^1, \dots, f_\theta^K\}$. Let $\sigma(\mathcal{A}, f_\theta(\mathcal{Q}, \mathcal{S})) \in \{0, 1\}$ be an indicator variable, where $\sigma = 1$ if $f_\theta$ correctly answers sample $x$, and $\sigma = 0$ otherwise. We define the empirical hardness score of sample $(\mathcal{Q}, \mathcal{S}, \mathcal{A})$ as the proportion of  $f_\theta$ in $\mathcal{F}$ that fail to answer it correctly:
\begin{equation}
    \mathcal{H}(\mathcal{Q}, \mathcal{S}, \mathcal{A}) = \frac{1}{K} \sum_{k=1}^K (1 - \sigma(\mathcal{A}, f_\theta^k(\mathcal{Q}, \mathcal{S}))
\end{equation}
Samples with a high empirical hardness score $\mathcal{H}(x)$ were identified as candidates for \textsc{OneEval}-Hard.

Subsequently, these candidates underwent multiple rounds of expert screening and review. Human experts qualitatively analyzed the samples, prioritizing those that specifically demand sophisticated reasoning capabilities known to be challenging for current LLMs. This includes instances requiring multi-step logical deduction, the association of implicitly stated knowledge, and the synthesis of information from disparate sources (cross-domain knowledge integration).

The final \textsc{OneEval}-Hard subset consists of 1,285 samples selected through this rigorous filtering process. By focusing on samples that empirically challenge existing models and qualitatively require complex reasoning patterns, \textsc{OneEval}-Hard offers higher discriminative power and represents a more challenging testbed. It serves as a critical resource for pinpointing specific knowledge blind spots and reasoning bottlenecks in LLMs, thereby providing an important foundation for driving targeted model analysis, optimization, and capability enhancement.

\section{Experimental Setup}
During the evaluation phase of \textsc{OneEval}, the parameters $\theta$ of the target LLM $f_\theta$ are kept constant. Each test sample's prompt is generated by integrating the user input $\mathcal{Q}$ with the retrieved external knowledge set $\mathcal{S}$, which is converted into a textual format to be compatible with the input requirements of $f_\theta$. 
The full prompt construction details of each task are provided in the Appendix.

\noindent \textbf{External Knowledge Retrieval Paradigm.}
As retrieval capability is not the focus of \textsc{OneEval}, the external knowledge set $\mathcal{S}$ for each test sample is obtained through a standardized retrieval approach and remains consistent across different instances of $f_\theta$. Specifically, leveraging dense retrieval techniques, the core methodology involves ranking knowledge fragments—such as text paragraphs, code snippets, or subgraphs of triples—based on the similarity $\sigma(\mathcal{Q}, \mathcal{S})$ between the dense vector representation of $\mathcal{Q}$ and $\mathcal{S}$, $\sigma(\mathcal{Q}, \mathcal{S}) = \cos(\mathbf{q}, \mathbf{s})$. 
Notably, the retrieved knowledge context may contain noise, reflecting real-world scenarios and providing a robust testbed to evaluate the model's resilience to imperfect or redundant information.

\noindent \textbf{Large Language Models.}
We selected a diverse set of representative LLMs, encompassing both open-source and closed-source models, a wide range of parameter scales, and various technical approaches. The models include Llama3.1-8B~\cite{llama3.1}, Llama3.1-70B~\cite{llama3.1}, GLM4-9B~\cite{glm4}, Qwen2.5-7B~\cite{qwen2.5}, Qwen2.5-72B~\cite{qwen2.5}, QWQ-32B~\cite{qwq}, DeepSeek-V3~\cite{deepseekv3}, DeepSeek-R1~\cite{deepseekr1}, Llama4-Maverick~\cite{llama4}, GPT-4o\cite{gpt4o}, GPT-4.1~\cite{gpt4.1}, o1~\cite{o1}, o3~\cite{o3o4}, o4-mini~\cite{o3o4}, Doubao-pro~\cite{doubaopro}, Claude3.7-Sonnet~\cite{claude3.7}, Grok3~\cite{grok3}, and Gemini-2.5-Pro~\cite{gemini2.5}.

\noindent \textbf{Evaluation Metric.}
In \textsc{OneEval}, we employ multiple evaluation metrics tailored to different tasks, including Accuracy, F1 score, and ISM@1. Detailed metrics for each task are outlined in Table 2. To ensure a fair assessment of a model's overall performance across tasks, we follow established LLM benchmarks and define the Overall Score as the average of its scores across all evaluation datasets.

\section{Results \& Analysis}

\begin{table*} 
	\begin{center}
	{\caption{Overall Score (\%) of \textsc{OneEval} and \textsc{OneEval}-Hard.}\label{tab:overall_results}}
	\scalebox{0.85}{
		\begin{tabular}{lcccccc}
			\toprule
			\textbf{Models} &\textbf{Release Date} &\textbf{Size} & \textbf{R-LLM} & \textbf{\textsc{OneEval}}  & \textbf{\textsc{OneEval}-Hard} & $\Delta$ \textbf{Hard}  \\
            
			\cmidrule(lr){1-7} 
            \multicolumn{4}{l}{\textit{Open-Source LLMs}} &  & \\
            \cmidrule(lr){1-7} 
            Qwen2.5-7B &Sep 2024 &7B & &$33.3$ &$10.7$ & $-22.6$ \\
            Llama3.1-8B &Jul 2024 &8B & &$31.0$ &$10.3$ & $-20.7$ \\
            GLM4-9B &Jun 2024 &9B & &$39.7$ &$13.2$ & $-26.5$ \\
            QWQ-32B &Mar 2025 &32B &\checkmark &$44.9$ &$15.5$ & $-29.4$ \\
            Llama3.1-70B &Jul 2024 &70B & &$45.8$ &$15.9$ & $-29.9$ \\
            Qwen2.5-72B &Sep 2024 &72B & &$46.0$ &$16.1$ & $-29.9$ \\
            Llama4-Maverick &Apr 2025 &400B & &$\textbf{48.2}$ &$\textbf{21.0}$ & $-27.2$ \\
            DeepSeek-V3 &Dec 2024 &671B & &$44.6$ &$17.8$ & $-26.8$ \\
            DeepSeek-R1 &Jan 2025 &671B &\checkmark &$47.2$  &$17.2$ & $-30.0$ \\
            
            \cmidrule(lr){1-7} 
            \multicolumn{6}{l}{\textit{Proprietary LLMs}} \\
            \cmidrule(lr){1-7} 
            GPT-4o &May 2024 &- & &$46.1$ &$16.3$ & $-29.8$ \\
            GPT-4.1 &Apr 2025 &- & &$58.1$ &$24.7$ & $-33.4$ \\
            o1 &Dec 2024 &- &\checkmark & - &$22.2$ & - \\
            o3 &Apr 2025 &- &\checkmark & - &$\textbf{32.2}$ & - \\
            o4-mini &Apr 2025 &- &\checkmark &$\textbf{53.1}$ &$29.4$ & $-23.7 $ \\
            Doubao-pro &Jan 2025 &- & &$42.1$ &$13.0$ & $-29.1$ \\
            Claude3.7-Sonnet &May 2025 &- &\checkmark &$40.9$ &$15.2$ & $-25.7$ \\
            Grok3 &Feb 2025 &- &\checkmark &$51.0$ &$21.7$ & $-29.3$ \\
            Gemini-2.5-Pro &Mar 2025 &- &\checkmark &- & $22.8$ & - \\
        
            \bottomrule
		\end{tabular}
		}
	\end{center}
\end{table*}

\subsection{Overall results}

Table~\ref{tab:overall_results} presents the overall accuracy of a diverse set of open-source and proprietary LLMs on both the standard \textsc{OneEval} and \textsc{OneEval}-Hard benchmarks. 
Due to space limitations, detailed experimental results for each dataset can be found in the Appendix.
Among the open source models, Llama4-Maverick achieved the best results thanks to its hybrid expert architecture. Among the private models, o4-mini and o3 achieved the best results on full set and hard set, respectively. In general, as the model size increases, the inference effect gradually improves. All models experience a substantial performance degradation when faced with the more challenging subset. For instance, the highest overall score on \textsc{OneEval}-Hard among open-source models (Llama4-Maverick, 21.0\%) remains markedly lower than its standard \textsc{OneEval} performance (48.2\%), corresponding to a drop of 27.2 percentage points. Similarly, proprietary models such as GPT-4.1 and o3, although leading on \textsc{OneEval}-Hard (24.7\% and 32.2\% respectively), still demonstrate large absolute gaps compared to their performance on the easier benchmark. The pronounced and universal decline across all models—often exceeding 25 percentage points—underscores the significant unresolved challenges. Therefore, we conclude \textbf{Insight 1: Even current state-of-the-art reasoning LLMs exhibit significant limitations when tackling knowledge-intensive reasoning tasks.}

\subsection{Performance on different KBs}
\begin{figure*}
\centering
	\includegraphics[width=\textwidth]{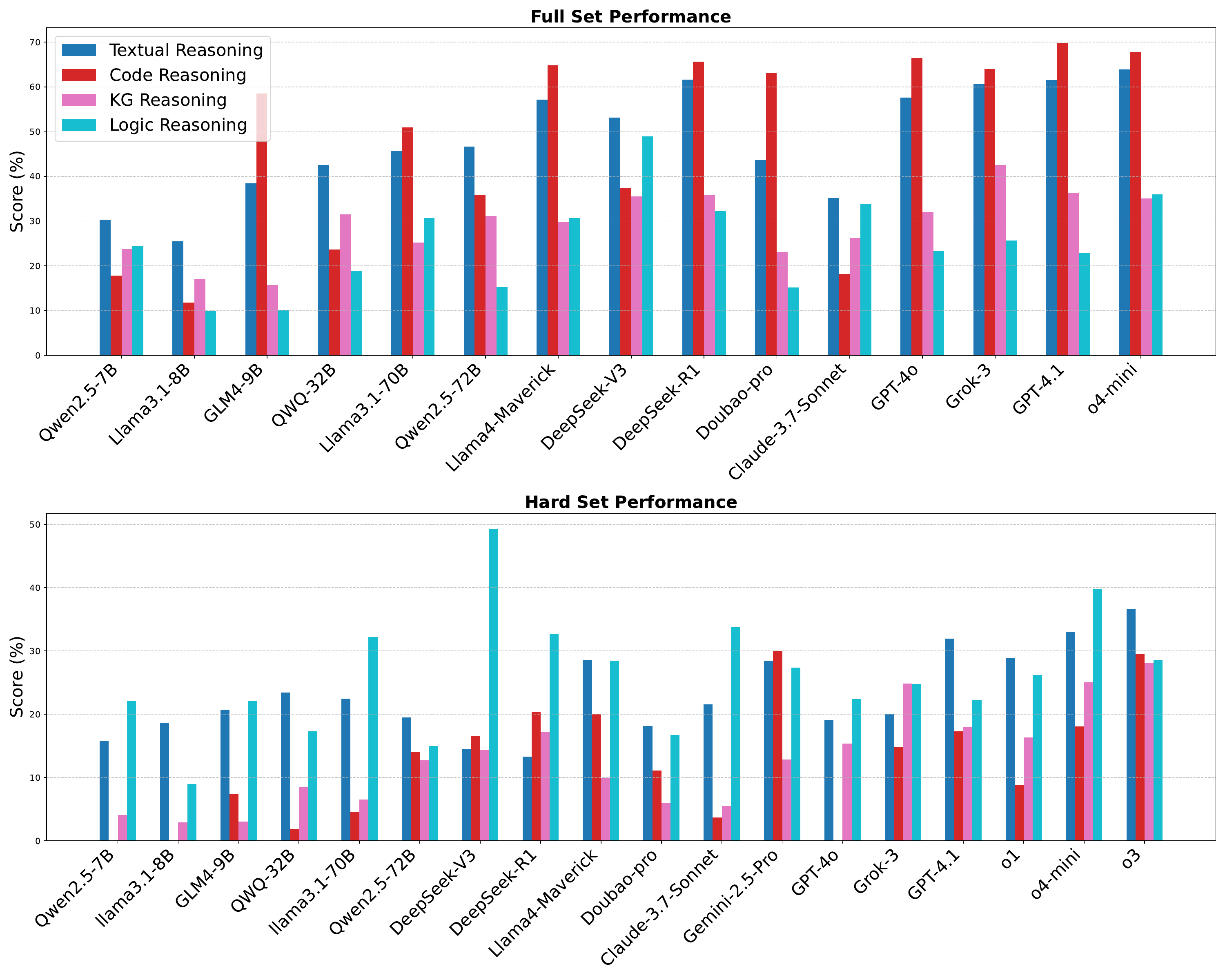}
	\caption{Performance across different knowledge bases for the Full set (top) and Hard set (bottom).} \label{fig:task_performance}
\end{figure*}

Figure~\ref{fig:task_performance} compares model performance across different knowledge bases for both the Full (top) and Hard (bottom) subsets of \textsc{OneEval}. Overall, Textual Reasoning tasks yield the highest scores, particularly for larger models such as GPT-4.0, GPT-4.1, and Q4-mini, suggesting that these tasks are relatively straightforward for current architectures. In contrast, Logic and Knowledge Graph (KG) Reasoning remain challenging, especially on the Hard subset, where all models exhibit pronounced performance drops.
Among code tasks, GPT-4.1 achieves the best results on the Full set, while Gemini-2.5-Pro and o3 show strong performance on the hardest instances. DeepSeek-V3 stands out on Logic Reasoning tasks, and Grok3 and o3—both designed for advanced reasoning—lead the KG benchmarks on the Full and Hard sets, respectively.
Furthermore, larger models consistently outperform smaller counterparts such as Qwen2.5-7B and Llama-3-8B, particularly in code and textual reasoning. However, even state-of-the-art models struggle with logic and KG tasks, underscoring current limitations in abstract and structured reasoning.



\subsection{LLM Performance with Increasing Knowledge Structuredness}
\begin{figure*}
\centering
	\includegraphics[width=\textwidth]{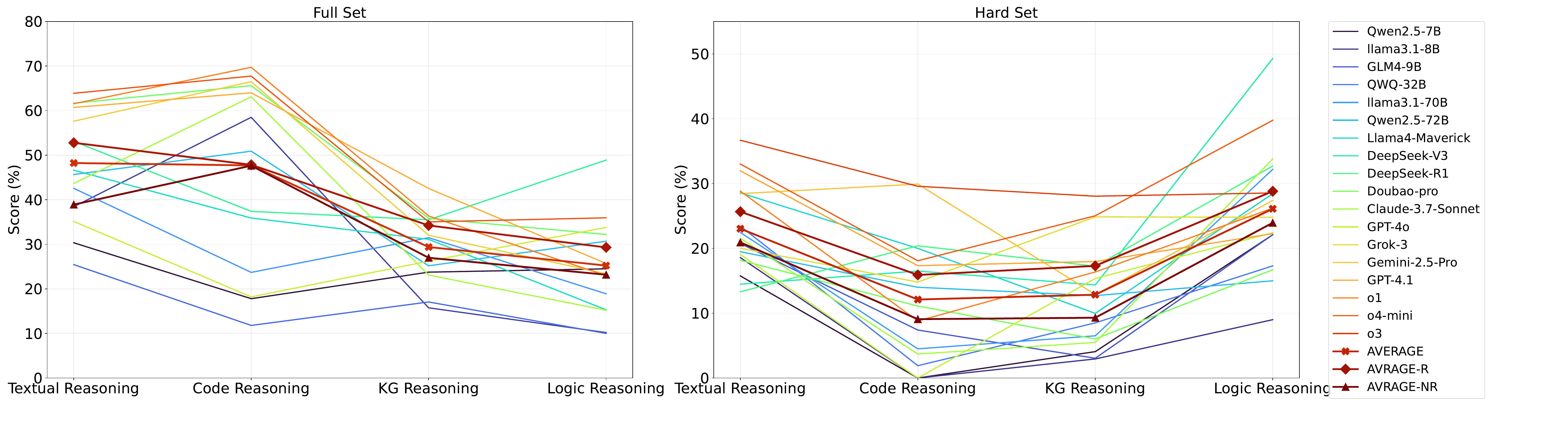}
	\caption{Score (\%) of \textsc{OneEval} and \textsc{OneEval}-Hard with Increasing Knowledge Structuredness.} \label{fig:task_level}
\end{figure*}

In our exploration of performance trends and rules governing  LLMs in reasoning tasks across different types of KBs, we organized the four KBs in \textsc{OneEval} by increasing levels of structuredness: text, code, knowledge graphs, and logic. We then analyzed the performance trends of various models on both the full and hard sets of these tasks. The experimental results are depicted in Figure~\ref{fig:task_level}. Here, AVERAGE, AVERAGE-R, and AVERAGE-NR represent the average of all model performances, the average of R-LLM performances, and the average of non-inferential LLM performances, respectively.

First, we point out \textbf{Insight 2: As the level of structure in tasks increases, the reasoning performance of LLMs tends to decline, indicating challenges in handling highly organized information.}. On the full set, the average score drops from 53 \% on textual reasoning to 25 \% on logic reasoning; on the hard subset the decline is even sharper (from 23 \% to 14 \%). The knee of the curve is the transition from code to KG reasoning: although most models clear 50 \% on code problems, they typically lose 25–30 absolute points when explicit graph navigation is required. This suggests that, while current instruction-tuned models have internalized a fair amount of “latent code semantics,” they still struggle to expose and manipulate discrete graph structures.

Second, models equipped with explicit inference scaffolds (“R-LLMs”) are systematically more resilient to increasing structure. The red curve (AVERAGE-R) lies above the brown curve (AVERAGE-NR) by 4.3 points on the full set and by 5.6 points on the hard set, with the gap widening for KG and logic reasoning. In other words, retrieval-augmented or step-by-step inference not only boosts raw accuracy, it scales better as symbolic demands grow. However, the advantage is not unlimited: the R-LLMs still trail human-level performance by a wide margin in logic, implying that retrieval alone is insufficient when deeper deductive chains are necessary.

Third, variance across models expands with structure. For text and code tasks the best and worst systems differ by about 20 points, whereas in logic reasoning the spread exceeds 30 points in both splits. Outliers are instructive: Grok-3 and Gemini 2.5-Pro maintain relatively high logic scores (32 \%), hinting that specialized training on formal proofs or chain-of-thought data can pay off. By contrast, smaller open-weight models like Llama-3-8B collapse to near-random on KG and logic problems, revealing a strong size–capability interaction for symbolic tasks.

\subsection{Cross-Modality Transferability}

\begin{wrapfigure}{r}{0.6\textwidth} 
\vspace{-10mm} 
	\includegraphics[width=0.6\textwidth]{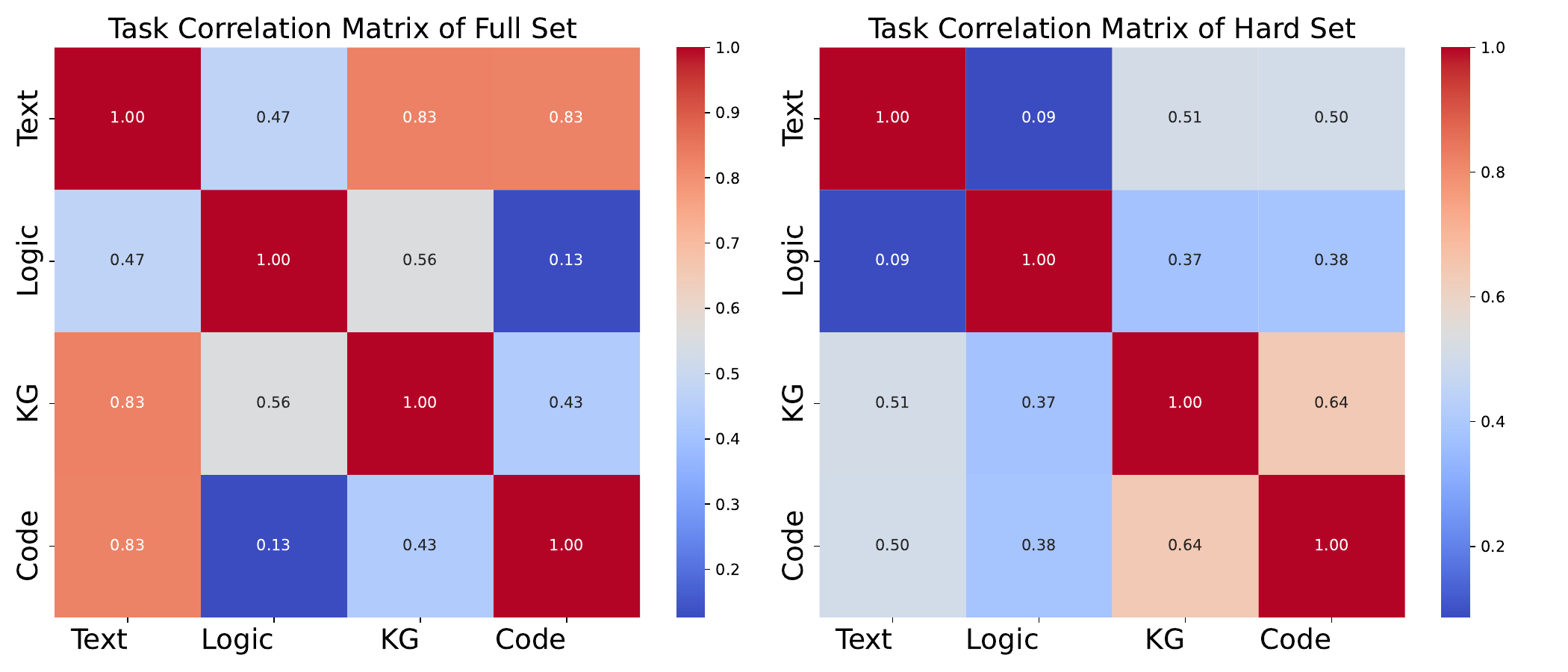}
	\caption{Correlation coefficients of performance on different tasks.} \label{fig:task_corr}
\vspace{-3mm} 
\end{wrapfigure}
We compute pairwise Spearman $\rho$ between model-level accuracies across four task families and observe a striking split (Figure~\ref{fig:task_corr}). On the full benchmark, Text shows uniformly high affinity with both KG ($\rho = 0.83$) and Code ($\rho = 0.83$), whereas Logic remains largely orthogonal ($\rho = 0.47$ to Text, $0.13$ to Code). The elevated Text $\leftrightarrow$ KG and Text $\leftrightarrow$ Code correlations indicate that, for standard-difficulty items, surface-level pattern matching learned from natural language suffices to solve many KG lookup and templated coding questions, yielding a shared representation space.

The picture changes on the hard subset, where spurious cues are removed. Text decouples almost entirely from Logic ($\rho = 0.09$) and only modestly aligns with KG/Code ($\approx 0.5$), while KG and Code remain mutually supportive ($\rho = 0.64$). This suggests two distinct reasoning substrates: (i) a symbolic substrate spanning KG and Code that survives distribution shift, and (ii) a shallow lexical substrate tying Text to others only when shortcuts are available. Practically, few-shot transfer from KG to Code should be effective even on adversarial data, whereas gains from Text pre-training will collapse under harder regimes—fine-tuning budgets should be allocated accordingly.

\subsection{Performance across Varing LLM Sizes}

\begin{figure*}
\centering
	\includegraphics[width=\textwidth]{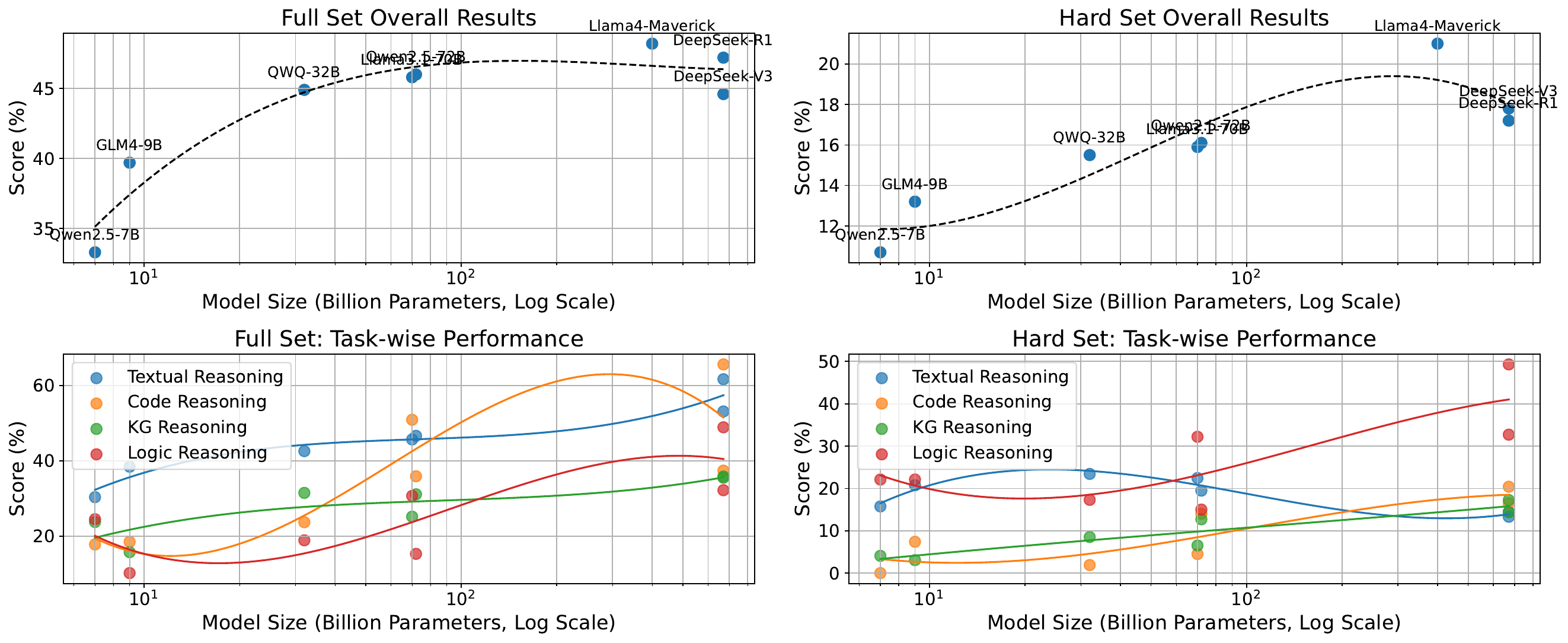}
	\caption{Reasoning performance for different model sizes.} \label{fig:scatter}
\end{figure*}

Experiments on our \textsc{OneEval} benchmark, designed to assess diverse reasoning capabilities, confirm established scaling laws~\cite{gpt4o}: LLM performance consistently improves with increased model size across both a Full set of tasks and a dedicated Hard subset. While larger models demonstrate significantly better overall reasoning scores, the Hard Set effectively probes the limits of current models, yielding substantially lower absolute scores and highlighting the inherent difficulty of these challenging reasoning instances. The trend suggests a potential plateau in performance for the very largest models evaluated, indicating diminishing returns on parameter scaling for these specific reasoning modalities with current architectures and training methodologies.

In the KG task, as the overall model size increases, the performance gap between models of the same size remains relatively unchanged. Conversely, in code-related tasks, this gap between same-sized models appears to widen as size increases.
It suggests that code tasks require more sensitive, emergent logical reasoning capabilities highly dependent on subtle model variations at scale, thus widening the performance gap between same-sized models. Conversely, KG tasks primarily leverage scaling for robust, structured knowledge processing, where within-size differences are more stable.

Most notably, Logic Reasoning shows limited positive scaling on the Full Set but demonstrates the most pronounced improvement with scale on the Hard Set. This suggests that larger LLMs possess enhanced logical processing capabilities that are primarily leveraged and become evident only when confronted with complex logical problems. The \textsc{OneEval} benchmark thus effectively differentiates models based on their size and architecture, providing critical insights into how various reasoning skills evolve with scale and identifying areas where current models still face significant limitations.

\subsection{Impact of LLM Thinking Length}

\begin{figure*}
\centering
	\includegraphics[width=0.5\textwidth]{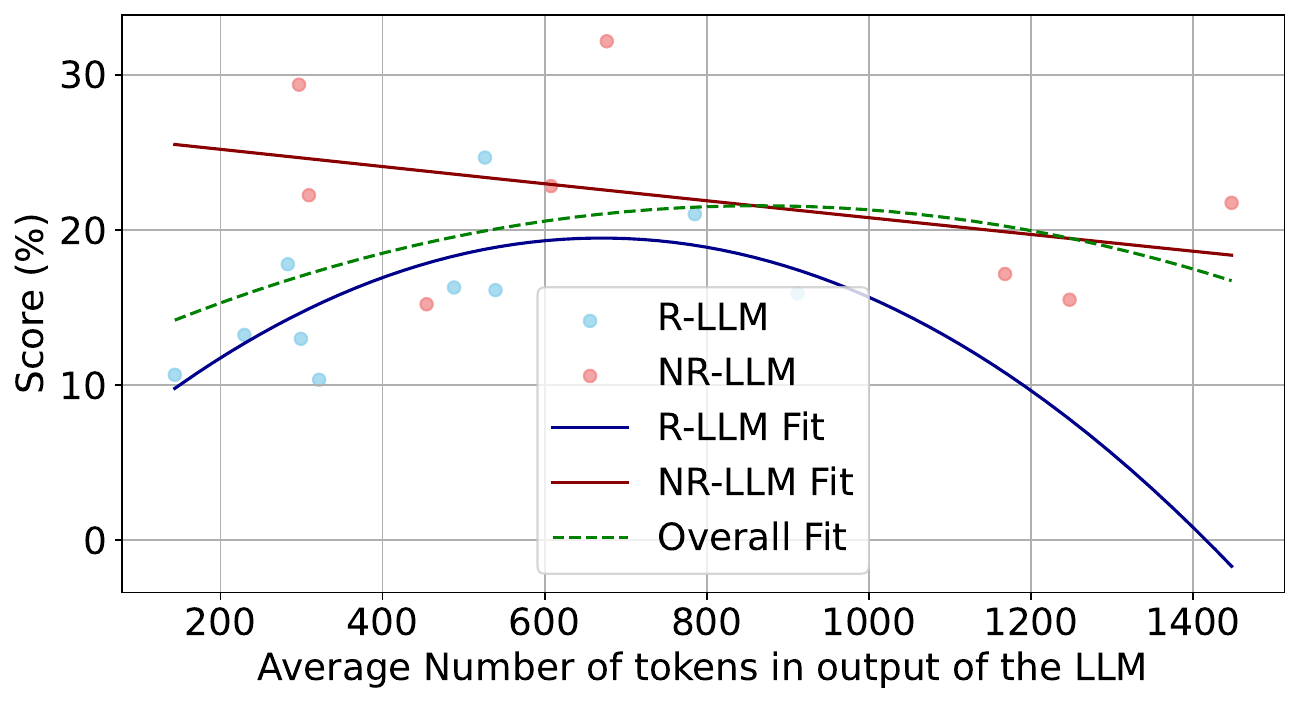}
	\caption{Model performance under different output lengths.} \label{fig:output_length}
\end{figure*}

Based on Figure~\ref{fig:output_length}, which plots LLM performance against output length (serving as a proxy for reasoning chain length), a clear distinction emerges between Reasoning-LLMs (R-LLM) and Non-Reasoning LLMs (NR-LLM). The fitted curve for R-LLMs exhibits a non-monotonic trend: performance initially increases with output length, reaching a peak at a moderate length, but then undergoes a significant decline as output length increases further. This empirical observation strongly supports the \textbf{Insight 3: Beyond a moderate length, the introduction of noise and potential errors outweighs the benefits of additional steps.}

The fitted curve for R-LLMs (blue line) exhibits a prominent non-monotonic relationship. Performance initially increases as output length grows from short chains (e.g., <400 tokens), indicating that some level of explicit reasoning is beneficial. The performance peaks at a moderate output length, roughly between 800 and 1000 tokens. Crucially, as the output length increases beyond this optimal point, the fitted curve shows a sharp decline in performance. For the longest output lengths depicted (>1200 tokens), the predicted performance drops significantly, even falling into negative score values.

The trend in R-LLMs suggests that while explicit reasoning steps are initially beneficial, there is an optimal chain length beyond which the accumulation of noise and potential errors with each additional step outweighs any potential gains. Unlike NR-LLMs, which show a more stable or gradually declining performance with length, R-LLMs demonstrate a vulnerability to excessive chain length, leading to performance degradation, which highlights the critical need for chain-length sensitivity and control mechanisms in complex reasoning tasks.

\section{Related Work}
\textbf{LLM Benchmarks.} Numerous benchmarks and leaderboards have explored a wide range of topics and tasks for evaluating diverse LLM capabilities. Common LLM benchmarks can be categorized into four types: (1) Knowledge Evaluation benchmarks test the LLM's mastery of subject knowledge through objective multiple-choice and open-ended questions, including MMLU~\cite{mmlu}, CMMLU~\cite{cmmlu}, CEval~\cite{ceval}, AGIEval~\cite{agieval}. (2) Instruction Following benchmarks like LLMBAR~\cite{llmbar}, Flan~\cite{flan}, and NaturalInstructions~\cite{naturalinstruction}, evaluate the LLM's ability to follow user instructions through QA formats. (3) Chat and Dialogue benchmarks focus on the LLM's contextual understanding and conversational abilities using multi-turn interactive question-answering data, including CoQA~\cite{coqa}, MMDialog~\cite{mmdialog}, MT-Bench~\cite{mtbench}, and OpenAssistant~\cite{openassistant}. (4) Safety and Risk benchmarks evaluate the LLM's safety and hallucination risks through multiple-choice and open-ended questions, such as DecodingTrust~\cite{decodingtrust}, AdvGLUE~\cite{advglue}, StrongReject~\cite{strongreject}, and HarmBench~\cite{harmbench}.

\textbf{LLM Leaderboards.} Unlike single evaluation benchmarks, the leaderboards integrate multiple evaluation tasks and consolidate a wide range of evaluation scenarios, adopting multi-dimensional and multi-task evaluation methods to construct a more comprehensive capability profile of LLMs. HuggingFace's OpenLLM\footnote{https://huggingface.co/open-llm-leaderboard} Leaderboard combines six benchmarks to comprehensively diagnosis LLMs' abilities in knowledge understanding, reasoning, mathematical problem-solving, information extraction, and complex task handling. Shanghai AI Lab's OpenCompass~\cite{opencompass} establishes an evaluation framework spanning six dimensions: examinations, knowledge, language, comprehension, reasoning, and safety, while incorporating dynamically updated evaluation datasets to ensure timeliness and breadth. FlagEval\footnote{https://flageval.baai.ac.cn} integrates over 30 capability dimensions and 30 benchmarks, constructing a large-scale evaluation repository with more than 100,000 test samples, covering multimodal information such as text, images, and audio. In addition to standardized dataset-based evaluations, comparative evaluations based on human preferences have gained significant attention in recent years. UC Berkeley's Chatbot Arena Leaderboard~\cite{chatbotarena} innovatively employs a crowdsourcing + Elo ranking mechanism, allowing human evaluators to pair-wise compare the quality of LLMs' responses, accumulating relative strength rankings among models. Stanford University's AlpacaEval~\cite{alpacaeval} introduces the ``LLM-as-a-Judge'' approach, leveraging large LLMs as judges to compare the responses of the evaluated model with those of a reference model, using relative win rates as the basis for ranking, significantly enhancing the scale and efficiency of evaluation.

However, existing benchmarks and leaderboards still have several limitations: (1) lack of evaluation for reasoning capabilities based on complex knowledge bases; and (2) homogeneous knowledge carriers, failing to comprehensively cover evaluating dimensions such as knowledge graphs, code, and structured tabular knowledge. To address these issues, \textsc{OneEval} introduces the first unified evaluation framework for cross-knowledge-source, multi-domain complex knowledge base reasoning tasks, aiming to achieve broader coverage and greater scientific rigor in evaluating the knowledge-enhanced capabilities of LLMs.

\section{Conclusion}

In this work, we introduced \textsc{OneEval}, a benchmark designed to evaluate LLMs on reasoning tasks involving structured external knowledge across various modalities and domains. Our evaluation of multiple state-of-the-art LLMs revealed significant limitations in structured reasoning, particularly as structural complexity increased. The models struggled with processing and reasoning over highly structured representations, and extending reasoning chains often led to diminishing returns. By releasing the \textsc{OneEval} dataset, evaluation scripts, and baseline results, we aim to provide a platform for advancing research in this area. While \textsc{OneEval} offers a diverse evaluation suite, its focus on static datasets may not fully capture the dynamic nature of real-world reasoning tasks, suggesting a direction for future improvements. Overall, our findings highlight the need for novel architectures, training paradigms, and reasoning strategies tailored to structured knowledge and formal systems.

{
\small
\bibliographystyle{unsrt}
\bibliography{neurips_2025}
}

\newpage
\appendix
\input{Appendix}

\end{document}

%% file: Appendix.tex

\definecolor{'wit'}{HTML}{FBFBFB}
\definecolor{'gry'}{HTML}{EEEEEE}

\definecolor{'deep1'}{HTML}{C5E6F8} 
\definecolor{'shallow1'}{HTML}{E4F3FC} 
\definecolor{'deep2'}{HTML}{E5F5B7} 
\definecolor{'shallow2'}{HTML}{F3FADF} 

\definecolor{'deep3'}{HTML}{FFE5C6} 
\definecolor{'shallow3'}{HTML}{FFF2E3} 
\definecolor{'deep4'}{HTML}{FFD3CF} 
\definecolor{'shallow4'}{HTML}{FFEAE8}
\definecolor{'deep5'}{HTML}{D2D0F3} 
\definecolor{'shallow5'}{HTML}{E8E7F9} 

\section{Details of Task Construction}

\subsection{BioTextQA}
\label{biotextqa}
In the pursuit of assessing the understanding and reasoning capabilities of large language models when applied to highly specialized scientific domains, we construct three QA datasets: BioTextQA, MatTextQA, and PharmKGQA. These datasets are designed to challenge models' understanding of biological sciences, material sciences, and pharmaceutical knowledge, respectively. The data sources for construction are derived from peer-reviewed publications and structured knowledge graphs within these domains: BioTextQA based on 3,881 biological papers from PubMed Central\footnote{https://pmc.ncbi.nlm.nih.gov/}, MatTextQA based on 8,222 materials science papers from arXiv\footnote{https://arxiv.org/}, PharmKGQA based on the PharmKG\footnote{https://github.com/biomed-AI/PharmKG} containing 1,093,237 pharmaceutical relational triples. The workflow for dataset construction is consistent across all three domains, summarized in three sequential steps as follows:

\textbf{\textit{Step 1. Knowledge Source Selection.}} Random samples of knowledge sources are extracted: for text-based domains like PubMed Central and materials science papers from arXiv, we randomly select entire documents or specific excerpts from the corpus. For PharmKG, relational triples conforming to predefined patterns are extracted.

\textbf{\textit{Step 2. Question Generation.}} The process for generating QA pairs differs slightly based on the type of knowledge source: text-based Knowledge Sources: We input the extracted textual content into GPT-4o, instructing it to generate questions directly related to the given text. This ensures that the questions are grounded in the provided knowledge source. KG-Based Sources: Questions are generated using predefined relational pattern templates, guaranteeing balanced distribution of question types and ensuring precise correspondence between questions and answers. Subsequently, GPT-4o is utilized to refine the questions, enhancing their semantic richness and fluency.

\textbf{\textit{Step 3. Question Validation.}} We employ GPT-4o for rigorous validation of generated QA pairs, focusing on two aspects: (1) meaningfulness of the question: The LLM assesses whether the question is coherent, contextually relevant, and scientifically valid;(2) answerability from the original knowledge source: LLM needs to verify whether the answer can be accurately traced back to the input text or knowledge graph. Questions must meet both requirements (both evaluated as ``True'') to be retained in the dataset. Through this systematic workflow, we ensure that the constructed datasets are of high quality—containing meaningful, relevant, and source-grounded QA pairs. These datasets serve as benchmarks for evaluating LLMs’ performance under the demanding criteria of scientific knowledge understanding and reasoning.

\subsection{MatTextQA}
The construction process of the MatTextQA dataset refers to the section~\ref{biotextqa}.

\subsection{ChineseLawFact}

ChineseLawFact is a fact-checking task that focuses on the Chinese legal domain and aims to verify the accuracy of legal statements. It consists of 9,464 annotated legal statements. Statements were extracted from objective questions in the Chinese National Judicial Examination, with corresponding explanations sourced from official exam preparation materials.  The task requires LLMs to possess not only deep knowledge of Chinese law, but also rigorous legal reasoning. 

To construct a high-quality fact-checking dataset, we utilized information from objective legal exam questions, including multiple-choice questions, their options, explanatory analyses, and associated legal knowledge points. When a question is deemed suitable for legal fact-checking tasks, we treat the question stem and each of the four answer options as four distinct factual claims. The accompanying explanatory analysis often contains legal provisions or knowledge that can serve as a basis for evaluating these claims—analogous to major premises in legal reasoning and these are extracted separately as part of the claim context. The remaining analysis can be divided into four segments, each corresponding to one of the answer options, providing correctness labels and detailed justifications. Each explanation serves as a minor premise in the structure of Chinese syllogistic legal reasoning, while the label represents the conclusion.

\subsection{AttributionNLI}
AttributionNLI is an automatically constructed benchmark designed to evaluate the fine-grained and logically complex attribution capabilities of Attributed Question Answering (AQA) systems. CAQA introduces a comprehensive attribution taxonomy, consisting of four categories: supportive, partially supportive, contradictory, and irrelevant. Moreover, it distinguishes between different levels of reasoning complexity in attribution: single, union, intersection, and concatenation. These dimensions allow CAQA to test a wide spectrum of attribution behaviors, from basic evidence recognition to complex multi-hop reasoning.

The construction of CAQA follows a four-step automated pipeline, grounded in knowledge graph (KG) semantics and logical query manipulation. The process is designed to scale easily across domains and supports the generation of high-quality attribution labels without manual intervention. 
(1) Query Collection.
We begin by collecting natural language questions, answer entities, and logical queries from existing KGQA datasets such as GrailQA and WebQuestionsSP. These logical queries, often expressed in SPARQL-like syntax, are aligned with Freebase, a structured knowledge graph comprising relational triples in the form (h, r, t). 
(2) Query Extension.
To simulate varying reasoning complexities, we apply logical operators to extend the original KG queries. Specifically, union and intersection operations are used to generate more complex query forms. These extensions transform single-triple, path-like, and tree-like queries into union-tree or intersection structures, which are associated with different complexity levels in attribution.
(3) Structured Attribution Generation.
Each extended query is grounded on the KG to extract a supporting subgraph. This subgraph is then programmatically edited to simulate other attribution categories:
Supportive: The original subgraph as retrieved.
Partially supportive: Derived by randomly deleting key triples from the subgraph.
Contradictory: Constructed by replacing the answer entity with a type-consistent but incorrect entity.
Irrelevant: Composed of unrelated triples while preserving structural similarity.
(4) Data Generation.
Using ChatGPT, we convert each subgraph into a natural language citation and paraphrase the question and answer accordingly. Each instance in CAQA thus contains a question, an answer statement, one or more citations, a categorical attribution label, and a complexity level.

\subsection{KCQAD}

Knowledge Conflict Question Answering Dataset is a novel resource comprising 500 carefully constructed questions designed to evaluate a model's ability to reconcile conflicting and evolving information. KCQAD captures both knowledge conflict and non-conflict scenarios, simulating real-world challenges where LLMs must navigate between their internal (parametric) knowledge and external (contextual) evidence to produce accurate answers. These scenarios are particularly relevant in dynamic domains such as current events, scientific discovery, and socio-political discourse, where information may change over time or differ across sources.

To systematically capture knowledge change, we focus on two primary phenomena: perpetuation and evolution. These subsets of KCQAD are automatically curated using structured temporal data extracted from Wikipedia tables and Wikidata triples. Specifically, we identify facts associated with time-indexed entities (e.g., political leaders, event outcomes, organizational status) and generate question-answer pairs reflecting different temporal states. For each pair, we retrieve context passages from the corresponding Wikipedia snapshots to provide supporting evidence. This design ensures that the dataset aligns with the natural evolution of knowledge, facilitating future extensions as new data becomes available.

In addition to temporal conflicts, KCQAD also incorporates misinformation-induced conflicts, where contradictions arise not from temporal updates but from deliberately incorrect information. To simulate these cases, we construct misinformation contexts by modifying original Wikipedia passages. This is done by replacing entities or facts with similar but incorrect alternatives in a way that maintains grammatical and semantic plausibility. These examples challenge models to detect and reason about factual inconsistencies that may not be apparent on the surface, closely mirroring the challenges posed by misinformation and fake news in real-world applications.

\subsection{PharmKGQA}
The construction process of the PharmKGQA dataset refers to the section~\ref{biotextqa}.

\subsection{AffairQA}
This task is formulated such that the input is a specific government affairs question $q$, and the output is a entity $e$ extracted from the government affair KG $\mathcal{K}$ as the answer of question. Its construction process can be divided into the following three steps: 

\textit{\textbf{Step1. Triple Extraction.}} The original government data is sourced from publicly available csv files on the Zhejiang Province data open platform\footnote{https://data.zjzwfw.gov.cn/jdop\_front/channal/data\_public.do}. We extract triples from csv files covering 5 different government scenarios to construct the government affairs KG $\mathcal{K}$. The document topics cover various topics, including city history, rescue experts, libraries, medical institutions, and gas stations information. 

\textit{\textbf{Step2. Data Cleaning and Validation.}} At this stage, we manually review whether there are any missing elements in these triples and filter out the unqualified ones. The final government affairs knowledge graph comprises 470,631 triples. 

\textit{\textbf{Step3. Question Generation.}} We upload the triples corresponding to each topic to the official Qwen website\footnote{https://chat.qwen.ai/}, prompting Qwen to randomly sample portions of the triples and synthesize corresponding questions. Human experts then perform a secondary verification of the generated questions. The resulting questions cover a variety of relation types, including but not limited to: containment relations, attribute relations, and spatiotemporal relations.

\subsection{PeopleRelQA}
People are among the core elements of human social activities and events. Therefore, information retrieval and reasoning-based question answering about people are very common. Since factual information related to individuals is typically stored and represented using structured knowledge graphs, we have constructed a complex people relationship-centric KGQA dataset PeopleRelQA. 

For knowledge base construction, we extract 785,553 people relationship-related knowledge triples from infoboxes in Baidu Baike\footnote{https://baike.baidu.com/} entries to serve as the KG source. Based on this KG $\mathcal{K}$, we design question templates based on three types of multi-hop relational patterns: multi-hop query (\texttt{A$\rightarrow$B$\rightarrow$C$\rightarrow$?}), path checking (\texttt{A$\rightarrow$B$\rightarrow$C$\rightarrow$D?}), and compositional query (\texttt{A$\rightarrow$B$\rightarrow$?$\leftarrow$D}). 

Furthermore, we utilize the graph query language to retrieve all paths from the people relationship KG $\mathcal{K}$ that fit these patterns and fill these paths into the question templates to obtain the initial question-answer pairs. To further increase question complexity, and ensure more natural expressions, we input these initial QA pairs into Qwen2-72B-Instruct for refinement, resulting in polished QA pairs involving multi-hop queries, counting tasks, logical compositions, and other complex KG reasoning questions.

\subsection{ReportFixer}
Financial reports sometimes contain inconsistencies or errors, which may result from data processing mistakes, subjective oversight, or asynchronous information updates. In practical applications, automatically identifying and correcting these inconsistencies enhances the reliability and credibility of the reports. This dataset is designed to evaluate the capability of large language model systems in detecting discrepancies between triples and text within financial reports. Consequently, the construction process of the ReportFixer dataset is as follows:

\textbf{\textit{Step 1. Data Collection.}} We search the CSCI Pengyuan website\footnote{https://www.cspengyuan.com/} using prefecture-level cities as keywords to find publicly available 2022 city investment company rating reports. In these financial reports, human experts manually select excerpts related to regional economic environments, including details such as economic development level, fiscal status, and debt level. Based on these selected texts, human experts compare the content of the selected excerpts with internal regional data (in the form of triples) from the company's reports. They also correct specific figures and remove data descriptions from the excerpts that are not present in the internal reports, thereby creating pairs of triples and text.

\textbf{\textit{Step 2. Hallucination Disturbance.}} We employ GPT-4o-0513 to introduce hallucination disturbances in the textual segments, generating naturally misleading text along with specific disturbance evidence.

\textbf{\textit{Step 3. Manual Annotation.}} Human experts use the specific disturbance evidence to compare the triples with the disturbed text, and identify all inconsistent triples. If the disturbance evidence in the text does not correspond to any triple facts, it is marked as none.

Based on the above process, we establish the ReportFixer task, whose input includes a set of triples $\mathcal{K}$ and a piece of disturbed text $\mathcal{D}$. The task requires identifying inconsistent triples within the input or confirming the consistency of the input information.

\subsection{VersiCode}

Existing LLMs struggle with library version dynamics in code generation. Based on this reason, we curate VersiCode, which was constructed through a multi-stage process involving data collection from three primary sources: popular Python libraries' source code (extracting API examples across versions), downstream application code from research papers (capturing real-world usage evolution), and Stack Overflow Q\&A pairs mentioning specific library versions. The raw data underwent rigorous cleaning using heuristic rules and hybrid human-LLM annotation to create structured metadata, followed by API lifecycle tagging to identify additions, deprecations, and version-specific behaviors. The final dataset was organized into task-specific subsets with executable test cases, incorporating version-controlled environments and comprehensive evaluation metrics to assess model performance on version-aware code generation and migration tasks. \\

\noindent\textbf{Task Type}
\begin{itemize}[leftmargin=*,nosep]
    \item Version-Specific Code Completion (VSCC)
    \item Version-Aware Code Migration (VACM)
\end{itemize}

\noindent\textbf{Data Collection:}
\begin{itemize}[leftmargin=*,nosep]
    \item Library Source Code: 300+ Python libraries (2,207 versions), API examples from docstrings
    \item Downstream Applications: Code from top-tier papers \& open-source projects
    \item Stack Overflow: Version-annotated Q\&A snippets
\end{itemize}

\noindent\textbf{Data Structure:} Each instance is structured as:
$$
m = \big[\text{Library } l;\ \text{Version } v;\ \text{Description } d;\ \text{Code } c\big]
$$
\begin{itemize}[leftmargin=*,nosep]
    \item \textbf{Lifecycle Tags}:
    \begin{itemize}[leftmargin=*,nosep]
        \item Addition: API introduced at $v_s$, active in $[v_s, v_e)$
        \item Deprecation: API deprecated at $v_e$ (last valid version)
    \end{itemize}
    \item \textbf{Granularities}: Token-level, line-level, \& block-level masking
\end{itemize}

\noindent\textbf{Migration Pairs:} Constructed as $(m_i, m_j)$ where:
$$
l_i = l_j,\quad d_i = d_j,\quad v_i \neq v_j
$$
covering both old$\rightarrow$new and new$\rightarrow$old migrations.

\noindent\textbf{Quality Control:}
\begin{itemize}[leftmargin=*,nosep]
    \item Hybrid Annotation: GPT-4 auto-labeling + human validation
    \item Metric: Critical Diff Check (CDC@k) for API usage, parameter matching, and syntax
\end{itemize}

\noindent\textbf{Statistics:}
\begin{itemize}[leftmargin=*,nosep]
    \item 11k+ completion instances, 76k+ migration instances
    \item Covers 9 years of library versions (2015--2023)
\end{itemize}

\subsection{SymenTex-ASC}
An ASP program is a set of rules of the following form:
\begin{align}
    \omega(\textbf{x}) \leftarrow 
    &\alpha_1(\textbf{x}_1), \ldots, \alpha_m(\textbf{x}_m),  \\
    &\text{not} \ \alpha_{m+1}(\textbf{x}_{m+1}), ... , \text{not} \ \alpha_n(\textbf{x}_n)
    \label{eq:definition_ASP_logic}
\end{align}
where each $\alpha_i(\textbf{x}_i)$ is a literal of the form $\texttt{p}(\textbf{x}_i)$ (positive literal) or $\texttt{-}\texttt{p}(\textbf{x}_i)$ (negative literal), and each $\textbf{x}_i$ consists of variables and constants.
In ASP, 
``\texttt{not}''  and ``\texttt{-}'' are called the default negation
and the classical negation (strong negation), respectively. An ASP program or a rule is ground if there are no variables.  A fact is a ground rule with $n=0$. We often write an ASP problem as a pair $(W,D)$ with $W$ a set of facts, and $D$ a set of rules.

For example, assuming the bird is named Tweety, the three ASP programs $P_i = (W_i, D), i =0,1,2$, where
\begin{equation}
\begin{aligned}
W_0 = \{& \text{Bird(\emph{Tweety})}\}
; W_1 = W_0 \cup \{\text{Injured(\emph{Tweety})}\} ;\\
& W_2 = W_1 \cup \{\text{SlightlyInjured(\emph{Tweety})} \}  \\
D = \{& \text{CanFly(A)} \leftarrow \text{Bird}(A), \text{not} \ \text{Abnormal(A)} ; \\
& \text{Abnormal(A)} \leftarrow  \text{Injured}(A), \text{not} \ \text{SlightlyInjured}(A)\} \label{eq:WD_example}
\end{aligned}
\end{equation}
Initially, since $W_0$ contains only ``Bird(\emph{Tweety})'', $P_0$ intuitively entails ``CanFly(\emph{Tweety})''. The new information ``Injured(\emph{Tweety})'' in $(W_1, D)$ triggers the second rule in $D$, entails ``Abnormal(\emph{Tweety})'', and invalidates the first rule in $D$. Finally the fact ``$\text{SlightlyInjured(\emph{Tweety})}$'' in $(W_2, D)$ invalidates ``Abnormal(\emph{Tweety})'', allowing ``CanFly(\emph{Tweety})'' to be inferred once again.

The semantics of ASP are characterized by the notion of answer sets, also known as stable models.
An answer set \emph{S} of (\emph{W}, \emph{D}) satisfies the following properties:

(1) $W \subseteq S$: All facts in \emph{W} are included in the answer set \emph{S}.

(2) For every rule $(\omega \leftarrow  \alpha_1, \ldots, \alpha_m,  \text{not} \alpha_{m+1}, ... , \text{not} \alpha_n) \in D$, if $\alpha_1, \ldots, \alpha_m \in S$ and $\alpha_{m+1}, ... , \alpha_n \notin S$, then $\omega \in S$. This ensures that the rules in \emph{D} are respected in \emph{S}.

\textbf{Answer Set Computation (ASC):} Given an ASP program $P$ (which is guaranteed to have one or more answer sets, potentially generated using disjunction in rule heads), the task is to compute and return one of its correct answer sets.

Formally, the task is to find a set $S'$ such that:
\begin{equation}
S' \in \text{AS}(P)
\end{equation}
The output is such a set $S'$. $\text{AS}(P)$ denote the set of all answer sets of the program $P$.

\section{Input and Output Examples for Each Task}

Table \ref{tab:instruction_biotextqa} shows the instruction and an example of BioTextQA.

Table \ref{tab:instruction_mattextqa} shows the instruction and an example of MatTextQA.

Table \ref{tab:instruction_ChineseLawFact} shows the instruction and an example of ChineseLawFact.

Table \ref{tab:instruction_AttributionNLI} shows the instruction and an example of AttributionNLI.

Table \ref{tab:instruction_kcqad} shows the instruction and an example of KCQAD.

Table \ref{tab:instruction_pharmkgqa} shows the instruction and an example of PharmKGQA.

Table \ref{tab:instruction_affairqa} shows the instruction and an example of AffairQA.

Table \ref{tab:instruction_peoplerelqa} shows the instruction and an example of PeopleRelQA.

Table \ref{tab:instruction_reportfixer} shows the instruction and an example of ReportFixer.

Table \ref{tab:instruction_versicode} shows the instruction and an example of VersiCode.

Table \ref{tab:instruction_symtex} shows the instruction and an example of SymTex-ASC.

\input{instruction/instruction-biotextqa}

\input{instruction/instruction-mattextqa}

\input{instruction/instruction-ChineseLawFact}

\input{instruction/instruction-AttributionNLI}

\input{instruction/instruction-kcqad}

\input{instruction/instruction-pharmkgqa}

\input{instruction/instruction-affairqa}

\input{instruction/instruction-peoplerelqa}

\input{instruction/instruction-reportfixer}

\input{instruction/instruction-versicode}

\input{instruction/instruction-symtex}

\section{Detailed Results on \textsc{OneEval}}
Table~\ref{tab:all_results} shows the performance of different LLMs on \textsc{OneEval}.
Table~\ref{tab:all_hardset_results} shows the performance of different LLMs on \textsc{OneEval}-Hard.

\begin{table}
\centering
\caption{Experimental results for each task in \textsc{OneEval}.}\label{tab:all_results}
\renewcommand{\arraystretch}{1.0}
\resizebox{\textwidth}{!}{\begin{tabular}{lcccccc}

\toprule
\textbf{Dataset}   & \textbf{AffairQA} & \textbf{BioTextQA} & \textbf{MatTextQA} & \textbf{PharmKGQA} & \textbf{ChineseLawFact} & \textbf{VersiCode}  \\
\textbf{Metric}        & \textbf{F1 (\%)}          &\textbf{F1 (\%)}          & \textbf{EM (\%)}      &\textbf{F1 (\%)}             &\textbf{ISM@1 (\%)}         & \textbf{F1 (\%)}      \\ 
\midrule
\multicolumn{7}{c}{\textit{Open Source LLMs}}\\ \midrule

Qwen2.5-7B       & 46.00  & 50.95   & 37.50   & 31.55   & 62.88        & 17.80   \\
Llama3.1-8B      & 42.00  & 55.23   & 55.98   & 23.53   & 57.13        & 11.80   \\
GLM4-9B          & 38.50  & 80.95   & 58.10   & 17.70   & 66.25        & 58.50   \\
QWQ-32B          & 45.00  & 76.67   & 62.38   & 45.67   & 69.00        & 23.70   \\
Llama3.1-70B     & 40.00  & 88.57   & 71.43   & 34.33   & 59.38        & 50.90   \\
Qwen2.5-72B      & 45.00  & 81.43   & 62.86   & 38.09   & 70.50        & 35.90   \\
Deepseek-V3      & 42.50  & 55.71   & 39.90   & 39.04   & 53.87        & 37.40   \\
DeepSeek-R1      & 45.50  & 33.81   & 50.48   & 31.37   & 58.00        & 65.60   \\
Llama4-Maverick  & 43.50  & 82.38   & 71.43   & 40.48   & 73.12        & 64.78   \\ 

\midrule
\multicolumn{7}{c}{\textit{Proprietary LLMs}}  \\ \midrule

GPT-4o           & 41.00  & 43.81   & 61.43   & 39.23   & 56.63        & 66.50   \\
GPT-4.1          & 47.00  & 46.67   & 69.05   & 40.95   & 73.62        & 69.72   \\
o4-mini          & 44.50  & 88.10   & 71.43   & 42.86   & 69.63        & 67.75   \\
Hunyuan-turbo    & 43.00  & 85.71   & 60.95   & 32.52   & 83.87        & 51.70   \\
Doubao-pro       & 40.00  & 83.33   & 50.00   & 27.14   & 57.50        & 63.10   \\
Grok 3           & 45.50  & 80.00   & 64.29   & 42.11   & 54.25        & 64.00   \\
Claude3.7-Sonnet & 22.00  & 78.10   & 48.80   & 40.10   & 60.38        & 18.20   \\

\toprule 
\textbf{Dataset}  & \textbf{PeopleRelQA} & \textbf{ReportFixer} & \textbf{KCQAD}   & \textbf{AttributionNLI} & \textbf{SymTex-ASC} &  \\
\textbf{Metric}        & \textbf{F1 (\%)}          &\textbf{F1 (\%)}          & \textbf{EM (\%)}      &\textbf{F1 (\%)}           & \textbf{EM (\%)}  &  \\ 

\midrule
\multicolumn{7}{c}{\textit{Open Source LLMs}}\\ \midrule

Qwen2.5-7B       & 0.50      & 17.00     & 42.00 & 35.80        & 24.50  &   \\
Llama3.1-8B      & 0.20      & 2.50      & 50.20 & 32.60        & 10.00  &  \\
GLM4-9B          & 0.20      & 6.60      & 55.00 & 45.10        & 10.20  &  \\
QWQ-32B          & 3.00      & 32.30     & 65.53 & 51.20        & 18.90   & \\
Llama3.1-70B     & 2.20      & 24.20     & 47.80 & 54.50        & 30.70  &  \\
Qwen2.5-72B      & 2.50      & 38.90     & 58.40 & 57.30        & 15.30  &  \\
Deepseek-V3      & 2.60      & 57.90     & 56.20 & 56.00        & 48.90  &  \\
DeepSeek-R1      & 6.80      & 59.70     & 64.44 & 71.20        & 32.20   & \\
Llama4-Maverick  & 3.50      & 32.10     & 36.20 & 52.00        & 30.67  &   \\ 

\midrule
\multicolumn{7}{c}{\textit{Proprietary LLMs}}\\ \midrule

GPT-4o           & 3.20      & 44.70     & 64.00 & 63.00        & 23.40  &  \\
GPT-4.1          & 4.00      & 53.50     & 55.60 & 55.60        & 22.97  &  \\
o4-mini          & 3.50      & 49.30     & 41.20 & 41.20        & 35.95  &  \\
Hunyuan-turbo    & 1.40      & 2.20      & 44.20 & 47.10        & 26.90  &  \\
Doubao-pro       & 0.00      & 25.30     & 41.08 & 60.10        & 15.20  &  \\
Grok 3           & 4.70      & 77.80     & 48.92 & 53.60        & 25.70  &  \\
Claude3.7-Sonnet & 0.50      & 42.30     & 43.30 & 62.70        & 33.80  &  \\
\bottomrule 

\end{tabular}}
\end{table}

\begin{table}
\centering
\caption{Experimental results for each task in \textsc{OneEval}-Hard.}\label{tab:all_hardset_results}
\renewcommand{\arraystretch}{1.0}
\resizebox{\textwidth}{!}{\begin{tabular}{lcccccc}

\toprule
\textbf{Dataset}    & \textbf{AffairQA} & \textbf{BioTextQA} & \textbf{MatTextQA} & \textbf{PharmKGQA} & \textbf{ChineseLawFact} & \textbf{VersiCode}  \\
\textbf{Metric}          & \textbf{Acc (\%)}          &\textbf{F1 (\%)}          & \textbf{EM (\%)}      &\textbf{F1 (\%)}             &\textbf{Acc (\%)}         & \textbf{ISM@1 (\%)}        \\ \midrule
\multicolumn{7}{c}{\textit{Open Source LLMs}} \\ \midrule

Qwen2.5-7B       & 0.00   & 26.67   & 17.79   & 14.29   & 21.67        & 0.00    \\
Baichuan2-7B     & 2.00   & 26.67   & 24.29   & 10.48   & 18.33        & 0.00    \\
Llama3.1-8B      & 0.00   & 28.10   & 26.79   & 11.76   & 31.67        & 0.00    \\
GLM4-9B          & 0.00   & 41.43   & 28.57   & 9.09    & 26.67        & 7.40    \\
Baichuan2-13B    & 0.00   & 29.05   & 10.95   & 7.62    & 40.00        & 1.90    \\
QWQ-32B          & 0.00   & 38.57   & 28.57   & 21.63   & 21.67        & 1.90    \\
Llama3.1-70B     & 0.00   & 44.29   & 35.24   & 15.42   & 20.00        & 4.50    \\
Qwen2.5-72B      & 0.00   & 40.95   & 31.90   & 21.43   & 6.67         & 14.00   \\
DeepSeek-V3      & 0.00   & 28.10   & 20.10   & 17.65   & 6.67         & 16.50   \\
DeepSeek-R1      & 0.00   & 9.52    & 1.43    & 5.39    & 21.67        & 20.40   \\
Llama4-Maverick  & 2.00   & 40.48   & 33.81   & 18.57   & 16.67        & 20.00   \\  

\midrule
\multicolumn{7}{c}{\textit{Proprietary LLMs}}\\ \midrule

GPT-4o           & 0.00   & 23.33   & 30.48   & 19.62   & 11.67        & 0.00    \\
GPT-4.1          & 2.00   & 24.29   & 35.71   & 18.57   & 23.33        & 17.33   \\
o1               & 0.00   & 42.45   & 33.65   & 21.15   & 18.33        & 8.80    \\
o3               & 2.00   & 41.90   & 38.46   & 22.49   & 23.33        & 29.58   \\
o4-mini          & 0.00   & 43.33   & 36.67   & 21.90   & 21.67        & 18.09   \\
Hunyuan-turbo    & 4.00   & 41.43   & 29.05   & 15.53   & 23.33        & 9.80    \\
Doubao-pro       & 0.00   & 42.38   & 22.86   & 11.90   & 16.67        & 11.10   \\
Claude3.7-Sonnet & 0.00   & 40.00   & 22.01   & 18.36   & 23.33        & 3.70    \\
Grok 3           & 3.00   & 40.00   & 29.52   & 19.62   & 11.67        & 14.80   \\
Gemini-2.5-pro   & 2.00   & 41.90   & 30.48   & 21.53   & 26.67        & 29.92   \\

\toprule 
\textbf{Dataset}    & \textbf{PeopleRelQA} & \textbf{ReportFixer} & \textbf{KCQAD}   & \textbf{AttributionNLI} & \textbf{SymTex-ASC}  \\
\textbf{Metric}            & \textbf{F1 (\%)}          &\textbf{F1 (\%)}          & \textbf{EM (\%)}      &\textbf{F1 (\%)}               & \textbf{EM (\%)}        \\ 
\midrule
\multicolumn{7}{c}{\textit{Open Source LLMs}}\\ \midrule

Qwen2.5-7B       & 2.00      & 0.00      & 2.20  & 10.40        & 22.10    \\
Baichuan2-7B     & 0.00      & 0.00      & 4.81  & 0.00         & 0.50     \\
Llama3.1-8B      & 0.00      & 0.00      & 4.20  & 2.10         & 9.00     \\
GLM4-9B          & 0.00      & 3.10      & 5.00  & 2.10         & 22.10    \\
Baichuan2-13B    & 0.00      & 0.00      & 3.40  & 0.00         & 1.00     \\
QWQ-32B          & 0.40      & 12.00     & 11.62 & 16.70        & 17.30    \\
Llama3.1-70B     & 0.80      & 9.80      & 4.40  & 8.30         & 32.20    \\
Qwen2.5-72B      & 8.10      & 21.30     & 5.40  & 12.50        & 15.00    \\
DeepSeek-V3      & 2.00      & 37.80     & 5.00  & 12.50        & 49.30    \\
DeepSeek-R1      & 8.00      & 55.60     & 8.89  & 25.00        & 32.70    \\
Llama4-Maverick  & 6.00      & 13.30     & 39.32 & 12.50        & 28.43    \\
\midrule

\multicolumn{7}{c}{\textit{Proprietary LLMs}}                                                               \\ \midrule
GPT-4o           & 2.50      & 39.30     & 11.00 & 18.80        & 22.40    \\
GPT-4.1          & 4.00      & 47.30     & 51.46 & 25.00        & 22.28    \\
o1               & 8.00      & 36.30     & 22.60 & 27.10        & 26.20    \\
o3               & 20.00     & 67.70     & 56.82 & 22.91        & 28.55    \\
o4-mini          & 4.00      & 74.20     & 42.54 & 20.83        & 39.76    \\
Hunyuan-turbo    & 3.00      & 0.80      & 3.40  & 2.10         & 24.90    \\
Doubao-pro       & 0.00      & 12.20     & 2.61  & 6.30         & 16.70    \\
Claude3.7-Sonnet & 0.40      & 3.20      & 3.60  & 18.80        & 33.80    \\
Grok 3           & 6.00      & 70.80     & 4.33  & 14.60        & 24.80    \\
Gemini-2.5-pro   & 8.00      & 20.00     & 22.33 & 20.83        & 27.37    \\
\bottomrule

\bottomrule
\end{tabular}}
\end{table}

%% file: instruction/instruction-biotextqa.tex
    \small
    \begin{longtable}{p{\linewidth}}
        \toprule
         \cellcolor{'shallow2'} \textbf{\textsc{Input:}} \cc{Answer the question based on the document.}\\
         \cellcolor{'shallow2'} \cc{Context: ['A tight cascade of gene regulation during the lifecycle of the malaria parasite in human blood cells suggests new functions for many Plasmodium genes', 'Plasmodium chabaudi chabaudi is a malaria parasite of murine rodents. It has been widely used as a model to study various aspects of parasite biology and disease which are difficult to investigate using human malaria parasites. For instance, P. c. chabaudi is being used to study the genetic basis of drug resistance [1-4] and strain-specific immunity [5], because the execution and analysis of genetic crosses is relatively straightforward in this species [6]. The analysis of the genetic basis of aspects of malaria biology has been facilitated by recent developments in malaria genomics. Firstly, the Plasmodium falciparum genome has been fully sequenced and mapped [7] and there is also extensive sequence data now available for three of the four main malaria parasites of murine rodents [8]. Secondly, the degree of homology and conservation of gene synteny between the various species of malaria [4,9,10] allows the undertaking of comparative genomics and facilitates the elaboration of accurate genomic maps in these species.', 'These data do not support the hypothesis of parasite growth inhibition due to an inhibition of the parasite SPM synthase activity as was demonstrated for PPMP [19-22] : 1) The anti-Plasmodium activity of AD2646 does not correlate with its inhibitory activity on the SPM synthase. Although AD2646 and PPMP showed similar inhibitory activity on this enzymic activity in parasites in cultures, AD2646 is about 300 times more efficient in inhibiting parasite development than PPMP; 2) In contrast to PPMP which inhibits the parasite development preferentially and reversibly at the ring stage [19], AD2646 inhibited parasite development preferentially and irreversibly at the trophozoite stage (Figure 5); 3) Inhibition of the SPM synthase activity by PPMP is associated with an inhibition of the TVN formation [19-22]. This was not observed in the presence of AD2646 (Figure 7).', 'Figure 2 Monthly parasite and blood smear examination incidence patterns. Monthly parasite incidence patterns of P. falciparum and P. vivax malaria combined per 1000 population (red line on logarithmic scale), blood smears examined per 1000 population (black line on logarithmic scale), and percentage of blood smears positive for malaria (blue line) from January 1995 to October 2004 in Sri Lanka.', 'Within-host competition in P. chabaudi is now firmly established [8,15,16]. Evidence for competition between co-infecting genotypes in human malaria infections is necessarily indirect, but consistent with this [4]. In older children and adults, for example, parasite densities do not increase with increasing numbers of clones, thus indicating that parasite clones are not regulated independently [17]. Given this, and the high frequency of mixed infections in human malaria [1-3,18] often consisting of both resistant and sensitive genotypes [19], and the fact that genetic diversity can be altered by antimalaria prophylaxis [20], it is highly likely that competitive release of drug resistance also occurs in human malaria. Indeed, a recent study has already implicated release of within-host competition as a key-factor in the spread of drug resistance in Uganda [21].']}\\
         \cellcolor{'shallow2'}Please use the context to answer the following question. You should give your answer in following format: [!Format Start!]A[!Format End!]. If there are multiple answers, please List all the answers divided with a comma in the same format.\\
         \cellcolor{'shallow2'}Question: what is  the genomic activity of the malaria parasite in human blood cells. Options:\\
         \cellcolor{'shallow2'}A. Genomic Activity of the Parasite in Human Blood Cells\\
         \cellcolor{'shallow2'}B. Microarray Analysis: Genome-Scale Data Gathering\\
         \cellcolor{'shallow2'}C. Genomic Activity of the Parasite in Mosquito Cells\\
         \cellcolor{'shallow2'}D. Parasitic Interaction with Red Blood Cells \\
         \cellcolor{'shallow2'}Let's think step by step! \\
        \midrule
        \vspace{-1mm}
        \cellcolor{'shallow2'} \textbf{\textsc{Output:}} \cc{A. Genomic Activity of the Parasite in Human Blood Cells}\\
        \bottomrule
    
    \caption{
  The instruction and an example of BioTextQA.}
    \label{tab:instruction_biotextqa}
    \end{longtable}

%% file: instruction/instruction-mattextqa.tex

    \small
    \begin{longtable}{p{\linewidth}}
        \toprule
         \cellcolor{'shallow2'} \textbf{\textsc{Input:}}  Context: ['ice phases, in contrast to data-intensive ML approaches that need the preparation of large trajectory datasets by running molecular dynamics simulations with carefully chosen force fields and/or pre-labeling efforts of water phases. 2) In the model-free classification step, instead of applying hand-crafted order parameters or machine-learned latent features, we employ a universal atomic descriptor, the Smooth Overlap of Atomic Positions (SOAP), 20 which ensures Euclidean symmetries with translation and rotation invariance and transferability to any structural system with different symmetries. This approach allows us to distinguish ice phases in the test datasets with a remarkable 100\% accuracy using only seven ideal reference structures of ice phases as model inputs. It demonstrates the generalizability of the score-based denoiser model in facilitating phase identification for complex molecular systems, and the unsupervised classification strategy proposed in this work can be generally applied to investigate structural evolution and phase identification for a variety of materials. Method Denoiser model As discussed in Ref., 19 the key component of the denoiser model is the noise prediction network    $\epsilon\theta(r)$ that predicts the “noise” or displacement vectors of input atomic structure r with respect to reference structure r0. Such a model can be used to “denoise” or remove', 'molecular dynamics\\ \cellcolor{'shallow2'} simulations, facilitating the subsequent classification tasks that identify phases by comparing the structural similarity 14 of denoised structures with ideal reference phases. The proposed approach achieves 100\% accuracy for phase classification of seven ice phases without the need for training trajectories and/or label information from MD simulations. The training of the denoiser model requires only ideal reference structures. The simplicity and generalizability of this framework allow it to be transferable and facilitate phase identification for complex polymorph structures. Moreover, the successful application of our denoising and classification approach to ice- liquid two-phase systems highlights its versatility in revealing the underlying crystalline structure of the ordered phases while preserving the liquid-like nature of the disordered phase. It showcases the possibility to enable a detailed analysis of thermodynamics and kinetics of the melting process, ice growth from super-cooled water, and the behavior of the ice/quasi-liquid interface at the atomic level without the need for prior phase labeling. This makes it a powerful tool for investigating complex systems where phase boundaries and transitions may not be well-defined or easily identifiable. In summary, our phase classification framework represents an important advancement in the accurate identification of ice phases in molecular dynamics simulations.', 'The gain in accuracy is significant by following our approach. In Figure 4, we show the results for two different models. The first model is using the approach laid out in Figure 1 and described in this paper. This results in a model with R2 = 86.6\%. To highlight the gain from our approach, we repeated the analysis following a traditional approach (labeled as “Processing as separate term”), where the annealing time and temperature were used as the two additional descriptors for the regression model, in addition to the parameters from the chemistry only network. Thus, time and temperature were used in place of the process descriptors discussed in Figure 3. This resulted in a model with the reduced accuracy of R2 = 69.1\%. Therefore, this added step to the framework resulted in a significant accuracy and therefore allows us to reasonably screen the entire chemistry / processing design space rapidly. We can now predict the capacity value for systems that have not been experimentally measured (ie. ‘virtual’ materials). We can randomly generate combinations of new chemistries and process conditions, and then repeat the process. While there are no required constraints in defining these compositions;']
         \\ \cellcolor{'shallow2'}Please use the context to answer the following question. You should give your answer directly in following format: [!Format Start!]Your Answer[!Format End!]. If there are multiple answers, please List all the answers divided with a comma in the same format.
         \\ \cellcolor{'shallow2'}Question: What is the the accuracy achieved for distinguishing ice phases using SOAP descriptor and ideal reference structures Options:
         \\ \cellcolor{'shallow2'}A. 95\% accuracy\\ \cellcolor{'shallow2'}B. 98\% accuracy\\ \cellcolor{'shallow2'}C. 85\% accuracy\\ \cellcolor{'shallow2'}D. 100\% accuracy  \cellcolor{'shallow2'} Let's think step by step! \\
        \midrule
        \vspace{-1mm}
        \cellcolor{'shallow2'} \textbf{\textsc{Output:}} \cc{C. 85\% accuracy}\\
        \bottomrule
    
    \caption{
  The instruction and an example of MatTextQA.}
    \label{tab:instruction_mattextqa}
    \end{longtable}

%% file: instruction/instruction-ChineseLawFact.tex
    \small
    \begin{longtable}{p{\linewidth}}
        \toprule

         \cellcolor{'shallow2'} \textbf{\textsc{Input:}} \cc{你是一名中经验丰富的中文法律专家，擅长法律事实核查验证，现在有一个情节和相关的法律声明，请根据专业知识判断其是否存在错误，并在最后输出结果`正确`或`错误`。

1.必要时，可以输出法条进行推理

2.提供详细的解释

3.一步步思考后给出结论

4.输出结果时请使用`结果`：`正确`或`错误`。

5.输出结果后，立即结束，不需要额外输出解释}\\
         \cellcolor{'shallow2'} \cc{\#\#\# 情节：`村集体雇了专业公司甲公司开飞机洒农药，飞机飞得低，且途经乙养鸡场。后乙养鸡场向丙履约，因为鸡的重量低于合同要求，损失10万元。乙养鸡场就认为是飞机把肉鸡吓得食欲下降饿瘦了，乙养鸡场和甲公司协商无果，将甲公司诉至法院。`}\\
         \cellcolor{'shallow2'}        \cc{\#\#\# 法律声明：`甲公司应当对没有因果关系承担责任`} \\
        \midrule
        \vspace{-1mm}
        \cellcolor{'shallow2'} \textbf{\textsc{Output:}} \cc{正确}\\
        \bottomrule
    
    \caption{
  The instruction and an example of ChineseLawFact.
    }
    \label{tab:instruction_ChineseLawFact}
    \end{longtable}

%% file: instruction/instruction-AttributionNLI.tex
    \small
    \begin{longtable}{p{\linewidth}}
        \toprule
         \cellcolor{'shallow2'} \textbf{\textsc{Input:}} Your task is to evaluate the relationship between a provided citation and the answer to a specific question. There are four possible types of relationships: \\ 
    \cellcolor{'shallow2'}Supportive: Choose this if the citation directly confirms or is fully in alignment with the answer, providing all necessary information to substantiate it. \\
    \cellcolor{'shallow2'} Insufficient: Choose this when the citation provides only partial backing for the answer, lacking some essential details or evidence needed for full support. \\
    \cellcolor{'shallow2'} Contradictory: Choose this option if the citation is consistent with the intent of the question but directly opposes or contradicts the answer. \\
    \cellcolor{'shallow2'} Irrelevant: Select this option if the citation does not match the intent of the question and contains information that is not useful for answering. \\
    \cellcolor{'shallow2'} For each example provided: First, you need to look at the question given and the answer provided. Then, compare them with the content of the citation. Finally, select the appropriate relationship category based on whether the citation supports the answer, is missing information, contradicts itself, or is irrelevant to the answer. \\
    \cellcolor{'shallow2'} Example: \\
    \cellcolor{'shallow2'} Question: what cheese made from milk of dromedary camel has the same texture as affidelice au chablis does?
    Answer: Caravane cheese is the cheese made from the milk of dromedary camel that has the same texture as affidelice au chablis. \\
    \cellcolor{'shallow2'} Reference: Caravane cheese is a type of food cheese that has a soft texture. It belongs to the category of soft cheeses along with Affidelice au Chablis.\\
        
        \midrule
        \vspace{-1mm}
        \cellcolor{'shallow2'} \textbf{\textsc{Output:}} \cc{Relationship Category: Insufficient}\\
        \bottomrule
    
    \caption{
  The instruction and an example of AttributionNLI.}
    \label{tab:instruction_AttributionNLI}
    \end{longtable}

%% file: instruction/instruction-kcqad.tex

    \small
    \begin{longtable}{p{\linewidth}}
        \toprule
         \cellcolor{'shallow2'} \textbf{\textsc{Input:}} \cc{Answer the question based on the document.}\\
         \cellcolor{'shallow2'} \cc{\#\#\#  Document: On 27 February 2024, Legends Z-A was announced during a Pokémon Presents presentation with a release window of 2025. At the end of the trailer, the Mega Evolution mechanic, which was first introduced in X and Y, was also teased to return. Pokémon Legends: Z-A is being developed by Electronic Arts. Pokémon Legends: Z-A is an upcoming 2025 video game developed by Electronic Arts and published by Nintendo and The Pokémon Company for the Nintendo Switch. Announced in February 2024, Legends: Z-A is part of the ninth generation of Pokémon video games. It takes place in Lumiose City in the Kalos region, based on Paris, France, which originated in Pokémon X and Y (2013). Legends: Z-A is the second Pokémon Legends game, following Legends: Arceus (2022). The game will take place in the Kalos region, which is based on France and was first introduced in the 2013 video games Pokémon X and Y. According to its trailer, it is set during an urban redevelopment project at Lumiose City, which is based on Paris, with Nintendo of America stating that it will take place entirely within Lumiose City.}\\
         \cellcolor{'shallow2'} \cc{\#\#\# Question: Who developed Pokémon Legends: Z-A?} \\
        \midrule
        \vspace{-1mm}
        \cellcolor{'shallow2'} \textbf{\textsc{Output:}} \cc{Game Freak.}\\
        \bottomrule
    
    \caption{
  The instruction and an example of KCQAD.
    }
    \label{tab:instruction_kcqad}
    \end{longtable}

%% file: instruction/instruction-pharmkgqa.tex

    \small
    \begin{longtable}{p{\linewidth}}
        \toprule
         \cellcolor{'shallow1'} \textbf{\textsc{Input:}} Triplets: [    `p105rb,gene,H,retinoblastoma,disease,``1844252.0, 7787878.0, 9546379.0, 7698220.0, 1844252.0, 2562189.0, 12210730.0, 7787878.0'',``'Because the product of the retinoblastoma gene , p105Rb , is expressed in all cell types , the obvious question is what accounts for these tissue specific differences in the role of p105Rb .', `Notably , both of the two giant regulators of checkpoint 1 -LRB- i.e. , p105RB -LSB- retinoblastoma oncosuppressor-encoded protein -RSB- and p53 dependent WAF1/CIP1 -RRB- are influenced by or influence G1 cyclins : cyclin E/cdk2 kinase complexes hyperphosphorylate p105RB , induce E2F release , and free G1 exit .', `The tumor suppressor genes p105RB -LRB- retinoblastoma , acting through the E2F transcription factor family -RRB- and p53 regulate cell proliferation , cell senescence , and apoptosis in many cell types .', `Substrates for cyclin\_D1 / Cdks have not been identified in vivo , but it has been proposed that the D class of cyclins might play a role in regulating the activity of the retinoblastoma gene product p105Rb .', `Because the product of the retinoblastoma gene , p105Rb , is expressed in all cell types , the obvious question is what accounts for these tissue specific differences in the role of p105Rb .', `Both the E1A and SV40 large T proteins contain similar CKII consensus sites proximal to the regions required for their associations with the retinoblastoma gene product -LRB- p105Rb -RRB- .', `The retinoblastoma susceptibility gene product , p105Rb -LRB- RB -RRB- , is generally believed to be an important regulator in the control of cell growth , differentiation , and apoptosis .', `Notably , both of the two giant regulators of checkpoint 1 -LRB- i.e. , p105RB -LSB- retinoblastoma oncosuppressor-encoded protein -RSB- and p53 dependent WAF1/CIP1 -RRB- are influenced by or influence G1 cyclins : cyclin E/cdk2 kinase complexes hyperphosphorylate p105RB , induce E2F release , and free G1 exit .' '']\\
         \cellcolor{'shallow1'}Please use the triplets to answer the following question. You should give your answer in following format: [!Format Start!]Answer[!Format End!]. If there are multiple answers, please List all the answers divided with a comma in the same format.\\
         \cellcolor{'shallow1'}Question: Please answer the following question: what entities are connected to 1 25 dihydroxycholecalciferol through O? Let's think step by step! \\
        \midrule
        \vspace{-1mm}
        \cellcolor{'shallow1'} \textbf{\textsc{Output:}} \cc{intercellular adhesion molecule 1}\\
        \bottomrule
    
    \caption{
  The instruction and an example of PharmKGQA.}
    \label{tab:instruction_pharmkgqa}
    \end{longtable}

%% file: instruction/instruction-affairqa.tex



    \small
    \begin{longtable}{p{\linewidth}}
        \toprule
         \cellcolor{'shallow1'} \textbf{\textsc{Input:}} \cc{根据以下三元组列表和您自己的知识背景，回答以下问题。在输出的最后一行，列出所有答案。你的答案应该仅包含用逗号分隔的答案。}\\
         \cellcolor{'shallow1'} \cc{\#\#\# 三元组信息: (永康历山省级森林公园，级别，省级），（永康历山省级森林公园，所在区县，永康市），（永康历山省级森林公园，所在城市，金华市），（永康历山省级森林公园，所在省，浙江省），（永嘉县五星潭省级森林公园，级别，省级），（永康历山省级森林公园，建造时间，2016/1/1），（永嘉县五星潭省级森林公园，所在区县，永嘉县），（永嘉县五星潭省级森林公园，所在省，浙江省），（永嘉县五星潭省级森林公园，所在城市，温州市），（遂昌县湖山森林公园有限公司，所在省，浙江省），（永康历山省级森林公园，type，公园），（铜铃山国家森林公园，地址，浙江省温州市文成县叶胜林场），（象山清风寨省级森林公园，级别，省级），（永嘉县五星潭省级森林公园，建造时间，2008/1/1），（诸暨杭坞山森林公园，所在省，浙江省），（象山南田岛省级森林公园，所在省，浙江省），（龙湾潭国家森林公园，所在区县，永嘉县），（牛头山国家森林公园，所在省，浙江省），（玉苍山国家森林公园，所在省，浙江省），（铜铃山国家森林公园，所在省，浙江省），}\\
         \cellcolor{'shallow1'}        \cc{\#\#\# 问题: 永康历山省级森林公园是在哪一年建造的？让我们一步一步思考！} \\
        \midrule
        \vspace{-1mm}
        \cellcolor{'shallow1'} \textbf{\textsc{Output:}} 2016/1/1\\
        \bottomrule
    
    \caption{
  The instruction and an example of AffairQA task.
    }
    \label{tab:instruction_affairqa}
    \end{longtable}

%% file: instruction/instruction-peoplerelqa.tex


    \small
    \begin{longtable}{p{\linewidth}}
        \toprule
         \cellcolor{'shallow1'} \textbf{\textsc{Input:}} \cc{根据以下三元组列表和您自己的知识背景，回答以下问题。在输出的最后一行，列出所有答案。你的答案应该仅包含用逗号分隔的答案。}\\
         \cellcolor{'shallow1'} \cc{\#\#\# 三元组信息:（胡适，主要作品，读梁漱溟先生的《东西文化及其哲学》），（拿什么拯救你，我的爱人，主要角色，祝四萍），（拿什么拯救你，我的爱人，主要角色，罗晶晶），（如果高中棒球的女经理人读过杜拉克的《管理学》的话，主要角色，朽木文明），（朱迪·福斯特，参演，魔幻迷宫：制作《沉默的羔羊》），（斯戴芬·莫昌特，参演，《临时演员》第一季），（斯戴芬·莫昌特，执导，《临时演员》第一季），（拿什么拯救你，我的爱人，主要演员，傅晶），（拿什么拯救你，我的爱人，主要演员，刘烨），（拿什么拯救你，我的爱人，主要演员，张谦），（拿什么拯救你，我的爱人，主要演员，于娜），（拿什么拯救你，我的爱人，主要角色，韩丁），（拿什么拯救你，我的爱人，主要演员，姚岗），（拿什么拯救你，我的爱人，主要角色，程瑶），（拿什么拯救你，我的爱人，主要角色，姚大维），（激流～你还记得我吗？～，主要角色，东萩耕司），（别担心，他不会走远的，主要演员，奥莉维亚·汉密尔顿），（妈妈，不当你的女儿可以吗？，主要角色，松岛太一），（维尼·琼斯，参演，《临时演员》第一季），（如果高中棒球的女经理人读过杜拉克的《管理学》的话，主要角色，宫田夕纪），（如果高中棒球的女经理人读过杜拉克的《管理学》的话，主要演员，峯岸南），（如果高中棒球的女经理人读过杜拉克的《管理学》的话，主要演员，铃木裕树），（巴黎，我爱你，主要演员，凯特琳娜·桑迪诺·莫雷诺），（如果高中棒球的女经理人读过杜拉克的《管理学》的话，主要角色，浅野庆一郎），（罗素·克劳，参演，《危情三日》的男人们），（对不起，我爱你，主要演员，沃拉甘·罗娜瓦查拉），（妈妈，不当你的女儿可以吗？，主要角色，早濑浩司），（别担心，他不会走远的，主要演员，凯瑞·布朗斯滕），（别担心，他不会走远的，主要演员，杰昆·菲尼克斯），（如果高中棒球的女经理人读过杜拉克的《管理学》的话，主要演员，大泉洋），（张雨生，音乐作品，爱的只是你（若我告诉你其实我爱的只是你）），（如果高中棒球的女经理人读过杜拉克的《管理学》的话，主要演员，濑户康史），（浪客剑心：传说的完结篇，主要角色，明神弥彦），（如果高中棒球的女经理人读过杜拉克的《管理学》的话，主要角色，北条文乃），（如果高中棒球的女经理人读过杜拉克的《管理学》的话，主要演员，石冢英彦），（如果高中棒球的女经理人读过杜拉克的《管理学》的话，主要角色，宫田靖代），（妈妈，不当你的女儿可以吗？，主要演员，南波瑠），（对不起，我爱你，主要演员，莫茶诺·欣彩萍翩），（丁丁历险记：独角兽号的秘密，主要角色，阿道克船长），（如果高中棒球的女经理人读过杜拉克的《管理学》的话，主要角色，川岛南），（如果高中棒球的女经理人读过杜拉克的《管理学》的话，主要角色，星出纯），（别担心，他不会走远的，主要演员，杰克·布莱克），（妈妈，不当你的女儿可以吗？，主要演员，麻生祐未），（拿什么拯救你，我的爱人，主要演员，李苒苒），（如果高中棒球的女经理人读过杜拉克的《管理学》的话，主要演员，池松壮亮），（纽约，我爱你，主要演员，卡洛斯·阿科斯塔），（激流～你还记得我吗？～，主要演员，南乃彩希），（如果高中棒球的女经理人读过杜拉克的《管理学》的话，主要演员，松岛庄汰），（黄磊，主要作品，我的肩膀，她们的翅膀），（别担心，他不会走远的，主要演员，鲁妮·玛拉）}\\
         \cellcolor{'shallow1'} \cc{\#\#\# 问题: 《利箭纵横》的主要演员的哪位搭档与陈道明搭档过？让我们一步一步思考！} \\
        \midrule
        \vspace{-1mm}
        \cellcolor{'shallow1'} \textbf{\textsc{Output:}} \cc{王志文}\\
        \bottomrule
    
    \caption{
  The instruction and an example of PeopleRelQA task.
    }
    \label{tab:instruction_peoplerelqa}
    \end{longtable}

%% file: instruction/instruction-reportfixer.tex



    \small
    \begin{longtable}{p{\linewidth}}
        \toprule
         \cellcolor{'shallow1'} \textbf{\textsc{Input:}} \cc{你是一位经济领域的专家，你将接收两个输入：1. 一组三元组，描述某个领域的事实。2. 一段描述相同或相关领域的文本。你的任务是判断这两种输入中描述的事实是否存在冲突。}\\
         \cellcolor{'shallow1'} \cc{三元组信息：``武进区'':[[2020, 综合财力, 478.75亿元], [2021, 转移性收入, 28.55亿元], [2020, 政府性基金收入, 247.86亿元], [2020, 转移性收入, 35.69亿元], [2021, 税收占比, 86.89\%], [2021, 政府性基金收入, 365.44亿元], [2021, 一般公共预算收入规模全市下辖区县中排名, 1], [2021, 综合财力, 610.55亿元], [2021, 税收收入, 188.16亿元], [2021, 一般公共预算收入增速, 10.93\%], [2021, 财政自给率, 103.42\%], [2021, 一般公共预算收入, 216.55亿元], [2021, 一般公共预算支出, 209.40亿元]]}\\ 
         \cellcolor{'shallow1'}
         \cc{文本：2021年武进区一般公共预算收入216.55亿元，较上年增长10.93\%，规模在常州市下辖区县中排名第1位；其中税收收入188.16亿元，税收占比86.89\%；一般公共预算支出209.40亿元，财政自给率103.42\%。政府性基金收入365.44亿元（上年同期为247.86亿元）；转移性收入28.55亿元（上年同期为35.69亿元）；综合财力为610.55亿元（上年同期为478.75亿元）。}\\
         \cellcolor{'shallow1'}
         \cc{请找出与文本不一致的三元组，这些三元组用逗号分隔，如果没有，请回答无。} \\
        \midrule
        \vspace{-1mm}
        \cellcolor{'shallow1'} \textbf{\textsc{Output:}} \cc{无}\\
        \bottomrule
    \caption{
  The instruction and an example of ReportFixer task.
    }
    \label{tab:instruction_reportfixer}
    \end{longtable}

%% file: instruction/instruction-versicode.tex
\small 
\begin{longtable}{p{\linewidth}}
\toprule
\cellcolor{'shallow4'}
\textbf{\textsc{Input:}} \\ \cellcolor{'shallow4'}
You are a professional Python engineer. \\ \cellcolor{'shallow4'}
I will provide functional descriptions and versions of specified dependency packages. \\ \cellcolor{'shallow4'}
You need to think step by step: 1. Understand the requirements 2. Consider how to use the dependencies 3. Implement the solution \\ \cellcolor{'shallow4'}
You need to write code in Python to implement this feature based on the functional description and using the dependency package and version I specified. \\ \cellcolor{'shallow4'}
Please note that you only need to return the code that implements the function, and do not return any other content. \\ \cellcolor{'shallow4'}
Please use \texttt{<start>} and \texttt{<end>} to enclose the generated code. Here is an example: \\ \cellcolor{'shallow4'}
\\
\cellcolor{'shallow4'}
\textbf{Function Description:} \\ \cellcolor{'shallow4'}
The function of this code is to print the results predicted by calling the model using vllm. \\
\cellcolor{'shallow4'}
\textbf{Dependency and version:} \\ \cellcolor{'shallow4'}
vllm==0.3.3 \\ \cellcolor{'shallow4'}
\textbf{Response:} \\ \cellcolor{'shallow4'}
\begin{verbatim}
<start>
for output in outputs:
    prompt = output.prompt
    generated_text = output.outputs[0].text
    print("Prompt,Generated text")
<end>
\end{verbatim}
\\
\cellcolor{'shallow4'}
\textbf{Dependency and version:} \\ \cellcolor{'shallow4'}
accelerate==0.20.0 \\
\cellcolor{'shallow4'}
\textbf{Functionality description of the code:} \\ \cellcolor{'shallow4'}
This code splits a list of elements between two processes and prints the inputs assigned to each process. The second split includes padding the inputs to make the assignment equal. \\
\cellcolor{'shallow4'}
\textbf{Refactored new code:} \\
\midrule
\vspace{-1mm}
\cellcolor{'shallow4'}
\textbf{\textsc{Output:}} \\
\cellcolor{'shallow4'}
\begin{verbatim}
<start>
# Assume there are two processes
from accelerate import PartialState
state = PartialState()
with state.split_between_processes(["A", "B", "C"]) as inputs:
    print(inputs)
# Process 0
["A", "B"]
# Process 1
["C"]
with state.split_between_processes(["A", "B", "C"], apply_padding=True) as inputs:
    print(inputs)
# Process 0
["A", "B"]
# Process 1
["C", "C"]
<end>
\end{verbatim}
\\ \bottomrule
\caption{The instruction and an example of VersiCode Task.}
\label{tab:instruction_versicode}
\end{longtable}

%% file: instruction/instruction-symtex.tex
\small 
\begin{longtable}{p{\linewidth}}
\toprule

\cellcolor{'shallow4'} \textbf{\textsc{Input:}} \\
\cellcolor{'shallow4'}\# Task: Generate a valid Answer Set (Stable Model) for the given ASP program. \\

\cellcolor{'shallow4'}\#\# Understanding the Task \\

\cellcolor{'shallow4'}You are given an Answer Set Programming (ASP) program consisting of facts and rules. Your task is to find *one* valid Answer Set (also known as a Stable Model) for this program. \\

\cellcolor{'shallow4'}\#\# Key ASP Concepts Recap \\

\cellcolor{'shallow4'}*   \textbf{Facts:} Ground atoms assumed to be true (e.g., `p(a).`). \\
\cellcolor{'shallow4'}*   \textbf{Rules:} Statements of the form `Head :- Body.` ("If `Body` is true, `Head` must be true"). \\
\cellcolor{'shallow4'}    *   `Body` can contain positive literals (`q(X)`), strongly negated literals (`-r(X)`), and default-negated literals (`not s(X)`). \\
\cellcolor{'shallow4'}    *   \textbf{Default Negation (`not`):} `not p` holds if `p` cannot be derived. \\
\cellcolor{'shallow4'}    *   \textbf{Strong Negation (`-`):} `-p` means `p` is explicitly false. \\
\cellcolor{'shallow4'}*   \textbf{Answer Set (Stable Model):} A set of ground literals `A` that is: \\
\cellcolor{'shallow4'}    1.  \textbf{Consistent:} Does not contain `p` and `-p` simultaneously. \\
\cellcolor{'shallow4'}    2.  \textbf{Stable:} `A` is the minimal classical model of the program's reduct `$P^A$` (formed by simplifying rules based on `A`). Essentially, everything in `A` must be derivable from the simplified program, and nothing more. \\

\cellcolor{'shallow4'} \#\# Goal \\

\cellcolor{'shallow4'} Your goal is to output a single set of ground literals that constitutes a valid Answer Set for the program defined by `[facts]` and `[rules]`. There might be multiple possible Answer Sets; you only need to provide one. \\

\cellcolor{'shallow4'}
\begin{verbatim}
<start>
- P17("Christopher").
- P18("John", "Charles", "Lucas").
- P19("John", "Charles", "Lucas").
- P20("John", "Charles").
- P21("Barbara", "John", "Lucas").
- P24("Christopher", "John", "Charles").
- P28("Christopher", "John", "Charles").
- P30("Barbara", "Christopher", "John").
P12("Barbara", "Christopher").
P13("Barbara", "John", "Charles").
P14("Barbara", "John", "Lucas").
P15("Christopher", "John", "Lucas").
P16("Christopher", "Charles", "Lucas").
P22("Barbara", "John", "Lucas").
P23("Barbara", "John", "Lucas").
P25("Christopher", "Charles").
P26.
P31("Barbara", "Christopher", "Lucas").
P32("Barbara", "Christopher", "Lucas").
- P0(V1, V3) :- P2(V1, V3, V4), not P5(V1, V3, V4), not -P16(V1, V3, V4).
- P1(V1) :- P0(V1, V3), P5(V1, V3, V4), P25(V1, V3).
- P11(V1, V3, V4) :- - P0(V1, V3), - P4(V0, V1, V3), - P6(V0, V1, V4).
- P12(V0, V1) :- P3(V0, V2, V4), P11(V1, V3, V4).
- P14(V0, V2, V4) :- P0(V1, V3), P3(V0, V2, V4), P6(V0, V1, V4).
- P29(V2, V4) :- - P30(V0, V1, V2), P31(V0, V1, V4), not -P32(V0, V1, V4).
- P3(V0, V2, V4) :- P12(V0, V1), - P20(V2, V3), - P21(V0, V2, V4).
- P4(V0, V1, V3) :- P13(V0, V2, V3), P14(V0, V2, V4), P15(V1, V2, V4).
- P5(V1, V3, V4) :- - P7(V2, V4), P10(V0, V1, V4), - P24(V1, V2, V3).
- P6(V1, V1, V1) :- P1(V1), not P8(V1), not P17(V1).
- P7(V2, V4) :- - P3(V0, V2, V4), not -P22(V0, V2, V4), not -P23(V0, V2, V4).
- P8(V0) :- - P3(V0, V2, V4), not -P22(V0, V2, V4), not -P23(V0, V2, V4).
- P9(V2, V0, V4) :- - P3(V0, V2, V4), not -P22(V0, V2, V4), not -P23(V0, V2, V4).
P1(V1) :- P2(V1, V3, V4), not P5(V1, V3, V4), not -P16(V1, V3, V4).
P10(V4, V4, V4) :- - P9(V2, V3, V4), not P18(V2, V3, V4), not P19(V2, V3, V4).
P11(V1, V3, V4) :- P8(V1), P9(V2, V3, V4), - P10(V0, V1, V4).
P11(V1, V4, V4) :- P3(V0, V2, V4), not -P6(V0, V1, V4), - P10(V0, V1, V4).
P18(V3, V3, V3) :- P0(V1, V3), not P1(V1), P8(V1).
P19(V1, V1, V1) :- - P1(V1), not -P26.
P2(V1, V3, V4) :- P12(V0, V1), - P20(V2, V3), - P21(V0, V2, V4).
P27(V1, V2, V3) :- - P28(V1, V2, V3).
P6(V0, V1, V4) :- P0(V1, V3), P3(V0, V2, V4), P11(V1, V3, V4).
<end>
\end{verbatim}
\\
\midrule
\cellcolor{'shallow4'} \textbf{\textsc{Output:}} \\
\cellcolor{'shallow4'} { -P17("Christopher"), -P18("John", "Charles", "Lucas"), -P19("John", "Charles", "Lucas"), -P20("John", "Charles"), -P21("Barbara", "John", "Lucas"), -P24("Christopher", "John", "Charles"), -P28("Christopher", "John", "Charles"), -P30("Barbara", "Christopher", "John"), P12("Barbara", "Christopher"), P13("Barbara", "John", "Charles"), P14("Barbara", "John", "Lucas"), P15("Christopher", "John", "Lucas"), {orange}{P16("Christopher", "Charles", "Lucas")}, P22("Barbara", "John", "Lucas"), P23("Barbara", "John", "Lucas"), P25("Christopher", "Charles"), P26, P31("Barbara", "Christopher", "Lucas"), P32("Barbara", "Christopher", "Lucas"), {P2("Christopher", "Charles", "Lucas")}, {P1("Christopher")}, P27("Christopher", "John", "Charles"), {-P0("Christopher", "Charles")}, {-P6("Christopher", "Christopher", "Christopher")}, -P3("Barbara", "John", "Lucas"), -P4("Barbara", "Christopher", "Charles"), -P29("John", "Lucas"), -P7("John", "Lucas"), -P8("Barbara"), -P9("John", "Barbara", "Lucas"), {P10("Lucas", "Lucas", "Lucas")} }

\\ \bottomrule
\caption{The instruction and an example of SymTex-ASC Task.}
\label{tab:instruction_symtex}
\end{longtable}